\documentclass[twoside,11pt]{article}
\usepackage[table,dvipsnames]{xcolor}
\usepackage{blindtext}

\usepackage[abbrvbib, preprint]{dmlr2e}

\usepackage{makecell}
\usepackage{multirow}
\usepackage{graphicx}
\usepackage{algorithm}
\usepackage{algorithmic}
\usepackage{amssymb,amsmath}
\usepackage{caption}
\usepackage{subcaption}

\usepackage{lastpage}
\usepackage{footnote}
\def\openreview{}

\firstpageno{1}
\usepackage{pifont}

\newcommand{\ourmethod}{VTruST}
\newcommand{\tb}[1]{\textcolor{blue}{#1}}

\algsetup{linenosize=\tiny}

\begin{document}

\title{VTruST: Controllable value function based subset selection for Data-Centric Trustworthy AI}

\author{\name Soumi Das\footnotemark[1]\ $^{12}$ \email soumidas@mpi-sws.org\\
       \name Shubhadip Nag\footnotemark[1] $^2$ \email shubhadipnag5555@gmail.com \\
       \name Shreyyash Sharma\footnotemark[1] \email shrysh1701@gmail.com \\
       \name Suparna Bhattacharya\footnotemark[2] \email suparna.bhattacharya@hpe.com \\
       \name Sourangshu Bhattacharya\footnotemark[1] \email sourangshu@cse.iitkgp.ac.in \\
       \AND
       \parbox{0.5\textwidth}{\addr\footnotemark[1] Dept. of Computer Science and Engg.\\
       Indian Institute of Technology, Kharagpur,\\
       Kharagpur, WB, India -721302.}\hfill
       \parbox{0.4\textwidth}{\addr \footnotemark[2] AI Research Lab \\    
       Hewlett Packard Labs, HPE \\ 
       Bengaluru, KA, India - 560048. }
       }


\maketitle

\begin{abstract}
Trustworthy AI is crucial to the widespread adoption of AI in high-stakes applications with \textit{fairness, robustness, and accuracy} being some of the key trustworthiness metrics. 
In this work, we propose a controllable framework for data-centric trustworthy AI (DCTAI)- VTruST, that allows users to control the trade-offs between the different trustworthiness metrics of the constructed training datasets. A key challenge in implementing an efficient DCTAI framework is to design an online value-function-based training data subset selection algorithm. We pose the training data valuation and subset selection problem as an online sparse approximation formulation.
We propose a novel online version of the Orthogonal Matching Pursuit (OMP) algorithm for solving this problem. 
Experimental results show that VTruST outperforms the state-of-the-art baselines on social, image, and scientific datasets. 
We also show that the data values generated by VTruST can provide effective data-centric explanations for different trustworthiness metrics.
\end{abstract}


\footnotetext[1]{Currently at: MPI-SWS, Saarbruecken, Germany.}
\footnotetext[2]{Equal contribution.}
\section{Introduction}





Trustworthiness \citep{kaur2022trustworthy,li2023trustworthy} of predictions made by Machine Learning models is crucial in many applications. In applications impacting society, e.g. loan eligibility prediction \citep{hardt2016equality}, criminal recidivism risk prediction \citep{Angwin2016Machine}, etc, fairness in prediction across different marginalized groups is as important as overall prediction accuracy. Similarly, the robustness of object detection systems for autonomous driving against perturbed input images\citep{song2024robustness}, or robustness against label corruption in phase transition prediction of sub-atomic particles \citep{Benato_2022} are important metrics compared to overall prediction accuracy. 
Tradeoffs between various notions of fairness with accuracy, e.g. individual fairness (demographic parity/equalized odds) \citep{roh2020fairbatch, romano2020achieving} or group fairness (Accurate Fairness) \citep{li2023accurate} are being studied for different models and training mechanisms. 
Similar studies have also been reported on inherent tradeoffs between feature robustness in images \citep{tsipras2018robustness,hu2023understanding} with accuracy, and adversarial label robustness \citep{pmlr-v162-pang22a,madry2018towards}.
While inherent tradeoffs between the trustworthiness metrics such as accuracy vs fairness or robustness is generally accepted, the nature of additional bias introduced by algorithms e.g adversarial training \citep{madry2018towards} or FairBatch \citep{roh2020fairbatch} is not clear. In this paper, we follow a data-centric approach to designing trustworthy AI techniques.

Data-centric approaches to AI model development \citep{zha2023data} strive to design methods for creating high-quality training datasets, that when used with standard SGD-based training algorithm can lead to models with specific trustworthiness properties. This eliminates the algorithmic bias introduced by specific algorithms, while limiting the bias only to the newly created training dataset, which is easier to interpret.
While many data valuation \citep{koh2017understanding,park2023trak,ghorbani2019data} techniques have been developed for the selection of high-quality training data subsets, most of them optimize only one property, e.g. validation set error rate. 
While well-known metrics for robustness and fairness exist, their use as a viable and efficient value function remains to be studied.
A key research issue in designing a data-centric approach is the design of an appropriate ``value function" that captures the notion of value of a training datapoint (toward trustworthiness metrics) while also being efficiently optimizable.
Another important research question is: can the value functions corresponding to various trustworthiness metrics be combined into a single value function using user-defined weightage? 
In this paper we address these research questions, effectively leading to a general data-centric framework to achieve user-controlled tradeoffs between different trustworthiness metrics. 

We propose additive value functions for accuracy, fairness, and robustness, which can be combined to form composite value functions.
The additiveness of the value functions is a key property that allows us to pose the problem of training data valuation and selection as an \textit{online sparse approximation problem}.
We propose a novel \textit{online orthogonal matching pursuit (OMP)} algorithm that greedily replaces features corresponding to a selected datapoint with those of a new datapoint, if there is a net improvement in the overall value of the selected set.
Unlike the traditional OMP \citep{5895106} which makes a pass through the entire training dataset to select an example, the proposed online OMP makes a  pass through the selected datapoints (a much smaller set) at the time of training update to optionally replace an existing selected point.
Experimental results on various applications demonstrate that models trained on subsets selected by VTruST can outperform all state-of-the-art baselines by $\sim 10-20\%$ and can also provide data-centric explanations behind its performance.

\newcommand{\cZ}{\mathcal{Z}}
\newcommand{\cD}{\mathcal{D}}
\newcommand{\cT}{\mathcal{T}}
\newcommand{\cS}{\mathcal{S}}
\newcommand{\cB}{\mathcal{B}}
\newcommand{\cA}{\mathcal{A}}
\newcommand{\cR}{\mathcal{R}}
\newcommand{\RR}{\mathbb{R}}
\newcommand{\cV}{\mathcal{V}}

\section{\ourmethod : Value-driven Trustworthy AI through Selection of Training Data}
\label{sec:method}

We propose a controllable value function-based framework for developing trustworthy models using a data-centric paradigm.
Our system has two components: (1) A general value function-based framework that allows users to specify a combination of trustworthiness metrics (sections \ref{sec:formulation} and \ref{sec:TAIvaluefn}), and (2) a novel online subset selection algorithm for constructing high-quality training dataset based on the specified value function (section \ref{sec:algo}). 

\subsection{A Controllable Value Function-based Framework for DCTAI}
\label{sec:formulation}

Let $\mathcal{D} = \{ d_i | i = 1,2,..,N\}$  be the training dataset and $\mathcal{D'} = \{ d'_j | j = 1,2,..,M\}$ be the validation dataset. Every datapoint $d'\in\cD'$ can be used to define the value function $\cV(\theta,d')$ which is used for calculating the value of a model $\theta$.
Given a run of model training, we define the incremental value function $v_i^t(d'_j)$ as the decrease in loss incurred due to an SGD update \citep{NEURIPS2020_e6385d39} using the datapoint $d_i$: $v_i^t(d'_j) =  l(\theta_t^{i-1},d'_j)-l(\theta_{t}^{i},d'_j)$, where $\theta_t^{i-1}$
and $\theta_t^i$ are the model parameters before and after the SGD update involving the training datapoint $d_i$ in the $t^{th}$ epoch.
Hence the value of a model $\theta^T$ can be defined as: $\cV(d'_j)= \sum_{t=1}^T \sum_{i=1}^N v^t_i(d'_j) \ \forall d' \in \cD'$.  We overload the notation to define the value function vector $\cV(\cD') = \sum_{t=1}^T \sum_{i=1}^N v_i^t(\cD')$, where $v_i^t(\cD') \in \RR^M$ is the vector of incremental values over all validation set datapoints.

Our data-centric framework aims to find a subset of training datapoints $\cS \in \cD$ that leads to a high-value model $\theta^t$ after training for $t$-epochs.
Let $\vec{y}_t = \sum_{k=1}^t \sum_{i=1}^N v^k_i(\cD')$ be the cumulative value function till the $t^{th}$ epoch.
We formulate the training data subset selection problem as a sparse approximation: $\vec{y}_t \approx \sum_{d_i\in S \subseteq \cD} \alpha_i [ \sum_{k=1}^t v_i^k (\cD') ] \ $, where $\alpha_i$ are the weights for the selected training datapoint $d_i$.
Next, using a second order Taylor series expansion of the change in loss function and plugging in the SGD update $\theta^i_t - \theta^{i-1}_t = \eta_t \nabla l(\theta^{i-1}_t,d_i)$, we obtain the following approximation for each term in the value function $ l(\theta_{t}^i,\mathcal{\cD'}) - l(\theta_{t}^{i-1},\mathcal{\cD'}) \approx \eta_t \nabla l(\theta_{t}^{i-1},d_i)^T  \nabla l(\theta_{t}^{i-1}, \mathcal{D'})+ \mathcal{O}(|| \theta_t^i - \theta_t^{i-1} ||_2^2)$. 
We truncate the Taylor expansion till the second-order terms
to arrive at the following sparse approximation problem:
   $ \vec{y}_t \approx \sum_{d_i\in S} \alpha_i \left[ \sum_{k=1}^t   \vec{X}_{i}^k \right]\ \forall t=1,...,T $
where $\vec{X}_{i}^k = \nabla l(\theta_{k}^{i-1},d_i)^T \nabla l(\theta_{k}^{i-1}, \cD') + \frac{(\nabla l(\theta_{k}^{i-1},d_i)^T \nabla l(\theta_{k}^{i-1}, \cD'))^2}{2}$ are the features for the $i^{th}$ training point calculated in epoch $t$. 
We use $\vec{y}_t$ and $\vec{X}_i^t$ to denote the predictor and predicted variables for valuating training datapoint $d_i$ using the entire validation set $\cD'$.
The main challenge in solving this approximation problem is that we need to store all the features $\vec{X}_{i}^k$ for all training datapoints $i$ over epochs $k=1,...,t$, in order to compute $ \sum_{k=1}^t \vec{X}_{i}^k$. This becomes prohibitively expensive. Instead, we solve the following \textit{online sparse approximation} (OSA) problem for each epoch $t$:
\begin{align}
    \vec{y}_t \approx \sum_{(p,q)\in S_t} \beta_p^q \vec{X}_{p}^q 
\end{align}
Here, $S_t$ is the set of selected training datapoints after epoch $t$. Note that the set $S_t$ can contain datapoints indexed by $p$ with features from any of the epochs $q=1,...,t$. We constrain the size of $S_t$ to be less than a user-specified parameter $\omega$. 
\sloppy
We describe an online algorithm for solving the above problem in Section \ref{sec:algo}.
Note that the value function $\cV(\cD')$ only needs to be additive over the training datapoints and epochs for the above formulation to be valid. Hence, this framework applies to a composite value function $\cV(\cD') = \sum_{f} \lambda_f \cV_f(\cD')$, where each value function $\cV_f(.)$ satisfies the additive property. This leads us to a general \textit{controllable} framework for incorporating many trustworthiness value functions, controlled using the user-specified weights $\lambda_f$.
\color{black}

\subsection{Value Functions for Trustworthy Data-centric AI}
\label{sec:TAIvaluefn}

For the accuracy metric, we use the value function proposed in \citep{NEURIPS2020_e6385d39}, which is defined as the decrease in loss incurred due to an SGD update using the datapoint $d_i$: $v_i^t(d'_j) =  l(\theta_t^{i-1},d'_j)-l(\theta_{t}^{i},d'_j)$ where $\theta_t^{i-1}$ and $\theta_t^i$ are the model parameters before and after the SGD update involving the training datapoint $d_i$ in the $t^{th}$ epoch.
Hence, the \textbf{accuracy value function} vector in defined as $\cV_a(\cD') = \sum_{t=1}^T \sum_{i=1}^N v_i^t(\cD')$.

\noindent
\textbf{Robustness Value Function}:
Training data augmentation using various perturbations has been observed to improve robust accuracy \citep{rebuffi2021data,addepalli2022efficient}. We use various perturbations to create the augmented training set $\cD_a$ and validation set $\cD'_a$ from $\cD$ and $\cD'$ respectively. The robustness value function is defined as $\cV_r(\cD'_a) =  \sum_{t=1}^T \sum_{d_i \in \{\cD \cup \cD_a\}} l(\theta_t^i,\cD'_a) - l(\theta_t^{i-1},\cD'_a)$. Since $\cV_r$ is derived from the loss function (that is additive), it also follows the additive property.

\noindent
\textbf{Fairness Value Function}: Existing literature in fairness \citep{roh2020fairbatch,romano2020achieving} uses equalized odds (EO) disparity and demographic parity disparity for achieving fair models. 
Let $x \in \mathcal{X}$ be the input domain, $\{y_0,y_1\} \in \mathcal{Y}$ be the true binary labels, and $\{z_0,z_1\} \in \mathcal{Z}$ be the sensitive binary attributes. We define the fairness value function as the change in EO disparity: $\cV_f(\cD') = \sum_{t=1}^T \sum_{d_i \in \cD} ed(\theta_t^i,\cD') - ed(\theta_t^{i-1},\cD')$. Based on \citep{DBLP:conf/iclr/Roh0WS21}, it is defined as the maximum difference in accuracy between the
sensitive groups ($z \in \mathcal{Z}$) pre-conditioned on the true label ($y \in \mathcal{Y}$): $ ed(\theta,\cD') = max(\|l(\theta , \cD'_{y_0,z_0}) - l(\theta , \cD'_{y_0,z_1})\| , \|l(\theta , \cD'_{y_1,z_0}) - l(\theta , \cD'_{y_1,z_1})\| )$. Considering we have $ed(\theta,\cD')$ defined for two validation sets $\cD'_1$ and $\cD'_2$ and the loss function is inherently additive, $ed(\theta,\cD'_1)+ed(\theta,\cD'_2) = ed(\theta,(\cD'_1+ \cD'_2))$ also holds true, thus $\cV_f$ turning out to be additive.

\noindent
\textbf{Composite value functions:} We can combine the value functions for accuracy ($\cV_a(\cD')$), robustness ($\cV_r(\cD'_a)$) and fairness ($\cV_f(\cD')$) to construct different composite value functions for observing tradeoffs between different trustworthiness metrics. We use the following combinations for our experiments: (a) Accuracy-Fairness : $\cV_{af}(\cD') = \lambda \cV_a(\cD') + (1-\lambda) \cV_f(\cD')$ ; (b) Accuracy-Robustness : $\cV_{ar}(\cD',\cD'_a) = \lambda \cV_a(\cD') + (1-\lambda) \cV_r(\cD'_a)$ ; (c) Robustness-Fairness : $\cV_{rf}(\cD',\cD'_a) = \lambda \cV_r(\cD'_a) + (1-\lambda) \cV_f(\cD')$.
The user-defined parameter $\lambda$ is used to control the tradeoff between the two objectives. 

\subsection{An online-OMP algorithm for online sparse approximation}
\label{sec:algo}

\begin{minipage}{0.46\textwidth}

\begin{algorithm}[H]

\tiny
\captionsetup{font=scriptsize}
 \caption{: \textbf{VTruST}}
 \label{algo:dataomp}
 \begin{algorithmic}[1]
 \tiny
 \STATE \textbf{Input:} \\
 i. $\omega$ : Total number of datapoints to be selected \\
 ii. $\vec{y}$ : Targeted value function \\
 iii. $\vec{X}_i$ : Features of all training points $d_i \in \cD$ \\
 iv. $S$ : Set of selected datapoint indices \\
 v. $\vec{\beta}\in \RR^{|S|}$: Weight of selected datapoints \\

 \STATE \textbf{Initialize:} \\
 $S \longleftarrow \phi$ \tb{//Indices of selected datapoints} \\ 
     \FOR{each epoch $t \in \{1,2,...,T\}$}
    \FOR{each datapoint $d_i \in \mathcal{D}$}
    \STATE \textbf{Input:} $\vec{y}_t$, $X_i^t \hspace{2mm} \forall i \in \{1,2,..,N\}, ||X_i^t||_2 = 1$
    \STATE \textbf{Process:}     
    \IF{$|S_{t-1}| = \omega$}
    \STATE $S_t \leftarrow$ \textbf{DataReplace}($\vec{y}_t, \vec{\xi}_{t-1}, S_{t-1}, \vec{\beta}_{t-1}, \vec{X}_i^t$)
    \ELSE
    \STATE $S_t \longleftarrow S_{t-1} \cup \{i\}$ \tb{// Add datapoints till the cardinality of $S_t$ reaches $\omega$}
    \ENDIF
    
 \STATE  Update $\vec{\beta}_t = \mbox{argmin}_\beta \|\vec{y}_t - \sum_{p,q \in S_t}(\beta_q^p \vec{X}_q^p) ||_2$
 
 \STATE Update $\vec{\xi}_t = \sum_{p,q\in S_t} \beta_q^p \vec{X}_q^p$
    \ENDFOR
    \ENDFOR
  \STATE \textbf{Output}:Final set of selected datapoint indices $S_T$, learned coefficients $\{ \beta_q^p| p,q\in S_T \}$
 \end{algorithmic}
\end{algorithm}
\end{minipage}
\hspace{5mm}
\begin{minipage}{0.46\textwidth}
\begin{algorithm}[H]
\captionsetup{font=scriptsize}
 \caption{:\textbf{DataReplace($\vec{y}_t, \vec{\xi}_{t-1}, S_{t-1}, \\ \vec{\beta}_{t-1}, \vec{X}_i^t$)} - \textit{Replace an existing datapoint}.}
 \label{algo:datareplace}
 \begin{algorithmic}[1]
 \scriptsize
 \STATE $\vec{\rho}_t = \vec{y}_t - \vec{\xi}_{t-1}$
 \STATE $\pi_{max}$ = -$\infty$
 \STATE $(a,b) = \phi$
\STATE  $\pi \longleftarrow \mbox{abs}(\vec{X}_i^t.\vec{\rho}_t)$
 \FOR{each index $p,q \in S_{t-1}$}
    \STATE $\pi'$ $\longleftarrow \mbox{abs}(\vec{X_q^p}.\vec{\rho}_t)$
    \STATE $\gamma$ $\longleftarrow \beta_q^p$
    \IF{$\pi > \pi'$ \& $\gamma \leq 0$ \& $(\pi' + \gamma) > \pi_{max}$}
    \STATE $\pi_{max} \longleftarrow \pi' + \gamma$
    \STATE $a,b \longleftarrow p,q $
    \ENDIF
\ENDFOR
\IF{$(a,b) \neq \phi$}
\STATE $S_t \longleftarrow S_{t-1} \setminus \{a,b\} \cup \{t,i\}$ 
\ENDIF
\STATE return $S_t$
\end{algorithmic}
\end{algorithm}
\end{minipage}
\normalsize

\noindent
In this section, we describe a novel online-OMP-based algorithm for the \textit{online sparse approximation problem}(OSA) in Algorithm \ref{algo:dataomp}.
The key difference between OSA and standard sparse approximation setting is that in OSA, new columns $\vec{X_p^q}$ are added and the target value $\vec{y_t}$ is updated at each epoch $t$. 
Line 10 in Algorithm \ref{algo:dataomp} adds new datapoints 
till the cardinality of $S_t$ reaches $\omega$. Once the buffer is saturated, the \textit{DataReplace} module 
is invoked in line 8 to replace an existing selected datapoint with the incoming datapoint. The criteria for replacement is to select the datapoints in $S_t$ that contribute to a better approximation of the current value function $\vec{y}_t$.
Hence a new datapoint with features $\vec{X_i^t}$ gets selected if the current approximation error reduces after the replacement. 
We compute the projection of the incoming datapoint features, $\vec{X}_i^t$ and that of the features of the selected datapoints $\vec{X}_q^p \mbox{ } \forall p,q \in S_t$ on the existing residual vector $\vec{\rho}_t$, measured by $\pi$ and $\pi'$ respectively.
We also denote by $\gamma$, the contribution of datapoint $p,q \in S_t$ obtained through $\beta_q^p$.
The datapoints with indices $(p,q)$ in $S_t$ whose additive impact ($\pi'+\gamma$) is smaller than that of incoming datapoint $(i,t)$, but larger than the current feature for replacement ($\vec{X}_q^p$) (line 8), gets substituted with the incoming point in line 14 of Algorithm \ref{algo:datareplace}. In terms of complexity, the per-epoch time complexity of OMP is $\mathcal{O}(\omega MN)$ and that of \textit{VTruST} is $\mathcal{O}(\omega M(N- \omega))$. 

\noindent
\textbf{Hyperparameter selection}: The proposed framework has two user-controlled hyperparameters, the tradeoff $\lambda$ and the subset-size $\omega$. Since the metrics are not monotone in $\omega$, we perform a grid search with various selection fractions between 10 - 90\%. Exploiting the monotonicity of metrics w.r.t. $\lambda$, one can fix a threshold on the first metric, say accuracy in case of $\cV_{af}$ and perform a binary search to arrive at an optimal point w.r.t. the second metric (fairness in $\cV_{af}$), once the threshold w.r.t. the first metric has been satisfied.

\color{black}
\section{Experimental Evaluation}

In this section, we describe the datasets, models, and evaluation metrics used for the trustworthiness metrics - Accuracy, Fairness and Robustness. 
We analyze the performance of \textit{VTruST} (VTruST-F with $V_{af}$, VTruST-R with $V_{ar}$, VTruST-FR with $V_{rf}$) over various applications. All our experiments have been executed on a single Tesla V100 GPU.

\subsection{Error rate, Fairness and Robustness on Social Data}

We evaluate the ability of VTruST to achieve a tradeoff between pairs of the three important social trustworthiness metrics: error rate (ER), fairness and robustness.
\color{black}
Our \textit{baselines} are: Wholedata standard training (ST), Random, SSFR \citep{roh2021sample}, FairMixup \citep{mroueh2021fair} and FairDummies \citep{romano2020achieving}. We report results on three benchmark datasets: COMPAS \citep{Angwin2016Machine} , Adult Census \citep{kohavi1996scaling} and MEPS-20 \citep{meps}.
We use a 2-layer neural network for all the datasets. We report two fairness metrics: equalised odds (EO) Disparity \citep{hardt2016equality} and demographic parity(DP) Disparity \citep{feldman2015certifying} following \citep{roh2021sample}.

\textbf{Fairness and Error Rate comparison (VTruST-F) with baselines}: 
We compare the performance metrics of VTruST-F with the baselines in Table \ref{tab:faireval}. 
The better the model is, the lower its ER as well as its fairness measures.
We can observe in Table \ref{tab:faireval} that VTruST-F with 60\% selected subset outperforms all the other methods in terms of fairness measures by a margin of $\sim 0.01-0.10$, and performs close to Wholedata-ST that yields the lowest ER. This denotes that it is able to condemn the error-fairness tradeoff emerging out to be the best performing method. We report these results with standard deviation across 3 runs. 

\begin{table*}[h]
\centering
\tiny
\captionsetup{font=scriptsize}
\caption{\textbf{Comparison of VTruST-F with baselines over 60\% subset for fairness evaluation.}}
\label{tab:faireval}
\begin{tabular}{|l|lll|lll|lll|}
\hline

\multicolumn{1}{|c|}{\multirow{2}{*}{\textbf{Methods}}}  & \multicolumn{3}{c|}{\textbf{COMPAS}}                                                                                         & \multicolumn{3}{c|}{\textbf{AdultCensus}}   & \multicolumn{3}{c|}{\textbf{MEPS20}}                                                                                    \\ \cline{2-10}  \multicolumn{1}{|c|}{}                                      
 & \multicolumn{1}{c|}{\textbf{\makecell{ER\\$\pm$std}}} & \multicolumn{1}{c|}{\textbf{\begin{tabular}[c]{@{}c@{}}EO\\ Disp \\ $\pm$std\end{tabular}}}  & \multicolumn{1}{c|}{\textbf{\begin{tabular}[c]{@{}c@{}}DP\\ Disp \\ $\pm$std\end{tabular}}} & \multicolumn{1}{c|} {\textbf{\makecell{ER\\$\pm$std}}} & \multicolumn{1}{c|}{\textbf{\begin{tabular}[c]{@{}c@{}}EO\\ Disp \\ $\pm$std\end{tabular}}} & \multicolumn{1}{c|}{\textbf{\begin{tabular}[c]{@{}c@{}}DP\\ Disp \\ $\pm$std\end{tabular}}} & \multicolumn{1}{c|}{\textbf{\makecell{ER\\$\pm$std}}} & \multicolumn{1}{c|}{\textbf{\begin{tabular}[c]{@{}c@{}}EO\\ Disp \\ $\pm$std\end{tabular}}} & \multicolumn{1}{c|}{\textbf{\begin{tabular}[c]{@{}c@{}}DP\\ Disp \\ $\pm$std\end{tabular}}} \\ \hline
\begin{tabular}[c]{@{}l@{}}\makecell{Wholedata-\\ ST}\end{tabular}    & \multicolumn{1}{l|}{\makecell{0.34\\$\pm$0.001}}       & \multicolumn{1}{l|}{\makecell{0.31\\$\pm$0.05}}           & \makecell{0.24\\$\pm$0.03}                                                               & \multicolumn{1}{l|}{\makecell{0.16\\$\pm$0.002}}       & \multicolumn{1}{l|}{\makecell{0.19\\$\pm$0.06}}           & \makecell{0.13\\$\pm$0.06}   &\multicolumn{1}{l|}{\makecell{0.09\\$\pm$ 0.001}}       & \multicolumn{1}{l|}{\makecell{0.09\\$\pm$ 0.007}}           & \makecell{0.08\\$\pm$ 0.0008}                                                                                     \\ \hline \begin{tabular}[c]{@{}l@{}}Random\end{tabular}  & \multicolumn{1}{l|}{\makecell{0.35\\$\pm$0.002}}       & \multicolumn{1}{l|}{\makecell{0.20\\$\pm$0.10}}           & \makecell{0.23\\$\pm$0.09}                                                                & \multicolumn{1}{l|}{\makecell{0.19\\$\pm$0.002}}       & \multicolumn{1}{l|}{\makecell{0.16\\$\pm$0.05}}           & \makecell{0.13\\$\pm$0.05}   &\multicolumn{1}{l|}{\makecell{0.12\\$\pm$0.017}}       & \multicolumn{1}{l|}{\makecell{0.06\\$\pm$0.02}}           & \makecell{0.08\\$\pm$0.005}                                                                                     \\ \hline

SSFR                                                        & \multicolumn{1}{l|}{\makecell{0.35\\$\pm$0.002}}    & \multicolumn{1}{l|}{\makecell{0.26\\$\pm$0.03}}               & \makecell{0.17\\$\pm$0.02}                                                                    & \multicolumn{1}{l|}{\makecell{0.21\\$\pm$0.001}}    & \multicolumn{1}{l|}{\makecell{0.18\\$\pm$0.03}}               & \makecell{0.12\\$\pm$0.01}   &\multicolumn{1}{l|}{\makecell{0.14\\$\pm$0.003}}       & \multicolumn{1}{l|}{\makecell{0.10\\$\pm$0.011}}           & \makecell{0.06\\$\pm$0.005}                         
\\ \hline
\begin{tabular}[c]{@{}l@{}}Fair-\\Dummies \end{tabular}     & \multicolumn{1}{l|}{\makecell{0.35\\$\pm$0.002}}     & \multicolumn{1}{l|}{\makecell{0.24\\$\pm$0.02}}              & \makecell{0.17\\$\pm$0.01}                                                                     & \multicolumn{1}{l|}{\makecell{0.16\\$\pm$0.002}}     & \multicolumn{1}{l|}{\makecell{0.14\\$\pm$0.01}}              & \makecell{0.10\\$\pm$0.01} &\multicolumn{1}{l|}{\makecell{0.12\\$\pm$0.001}}       & \multicolumn{1}{l|}{\makecell{0.13\\$\pm$0.005}}           & \makecell{0.08\\$\pm$0.003}                    

\\ \hline
\begin{tabular}[c]{@{}l@{}}Fair-\\Mixup \end{tabular}  & \multicolumn{1}{l|}{\makecell{0.35\\$\pm$0.03}}     & \multicolumn{1}{l|}{\makecell{0.15\\$\pm$0.03}}              & \makecell{0.13\\$\pm$0.04}                                                                     & \multicolumn{1}{l|}{\makecell{0.24\\$\pm$0.04}}     & \multicolumn{1}{l|}{\makecell{0.11\\$\pm$0.05}}              & \makecell{0.1\\$\pm$0.02} &\multicolumn{1}{l|}{\makecell{0.89\\$\pm$0.02}}       & \multicolumn{1}{l|}{\makecell{0.02\\$\pm$0.04}}           & \makecell{0.05\\$\pm$0.03}                 

\\ \hline 

\rowcolor{LimeGreen} VTruST-F                                                   & \multicolumn{1}{l|}{\makecell{0.34\\$\pm$0.002}}   & \multicolumn{1}{l|}{\textbf{\makecell{0.15\\$\pm$0.01}}}                & \textbf{\makecell{0.13\\$\pm$0.01}}                                                                   & \multicolumn{1}{l|}{\makecell{0.18\\$\pm$0.001}}   & \multicolumn{1}{l|}{\textbf{\makecell{0.11\\$\pm$0.03}}}                & \textbf{\makecell{0.05\\$\pm$0.01}}  &\multicolumn{1}{l|}{\makecell{0.09\\$\pm$0.003}}       & \multicolumn{1}{l|}{\textbf{\makecell{0.01\\$\pm$0.001}}}           & \textbf{\makecell{0.05\\$\pm$0.0008} }                                                                                   \\ \hline
\end{tabular}
\vspace{-5mm}
\end{table*}


\textbf{Tradeoffs between Error rate, Fairness and Robustness (VTrust-F , VTruST-FR)}:
We observe the tradeoffs between error rate vs fairness (VTruST-F: Figure 1a) and fairness vs robustness (VTruST-FR: Figure 1b) through pareto frontal curve by varying $\lambda$ $\in \{0,0.1,0.3,0.5,0.7,0.9,1\}$.
The error rate is measured on the clean test sets while robust error rates are measured on the label flipped test sets.
We can observe that Wholedata-ST has a lower error rate but high disparity and robust error values. The other baselines continue to have a higher error rate and disparity compared to VTruST. We report the results on other datasets in the Appendix.

\begin{figure*}[h!]

    \centering
    \begin{subfigure}{\textwidth}
    \centering
    \includegraphics[width=0.45\textwidth,height=2.5cm]
    {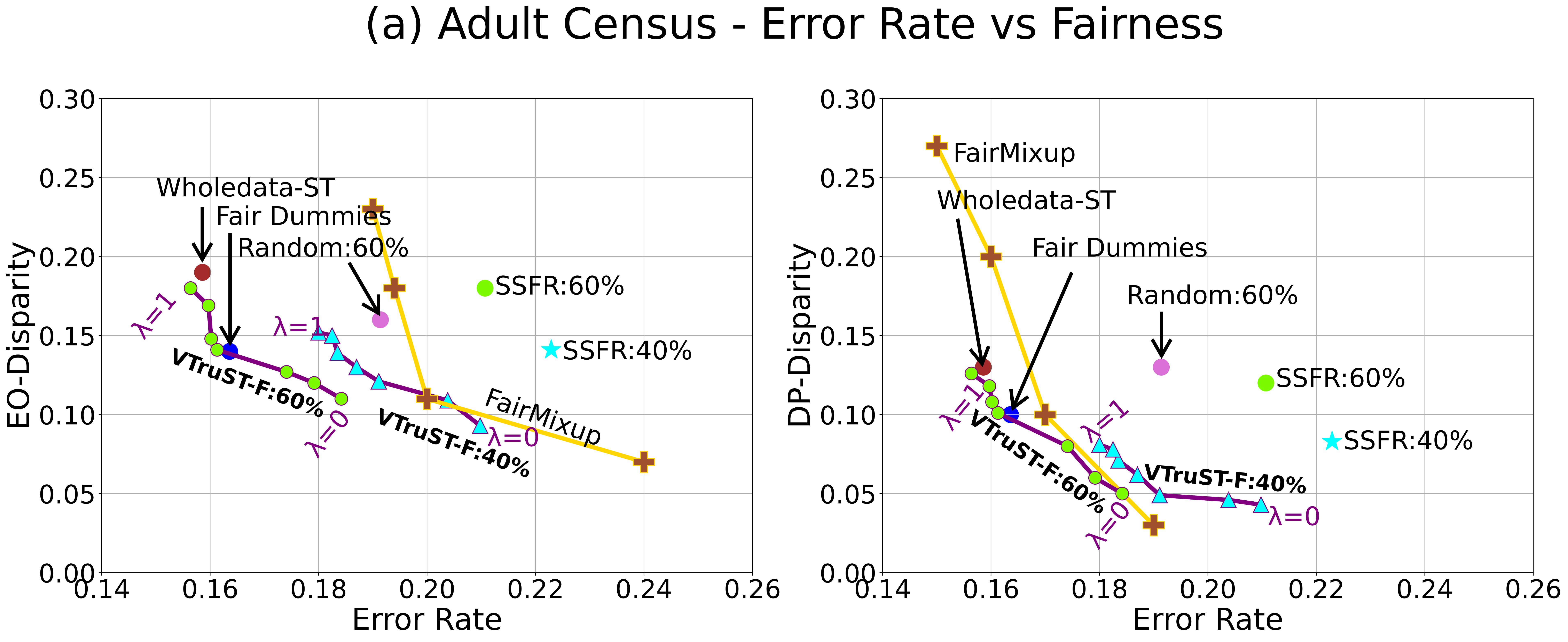}
    \end{subfigure}
    \begin{subfigure}{\textwidth}
    \centering
    \includegraphics[width=0.45\textwidth,height=2.5cm]{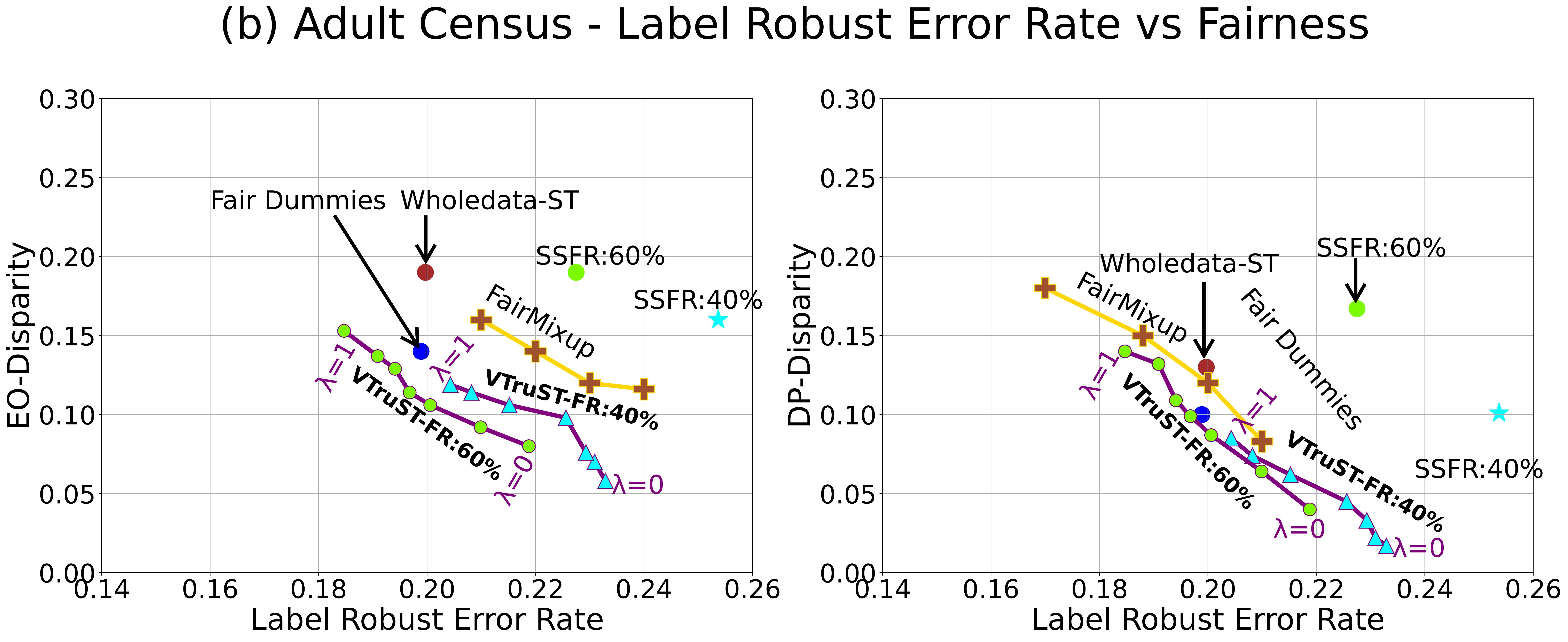}
    \end{subfigure}

    \captionsetup{font=scriptsize}
    \caption{\textbf{Controlling tradeoffs in trustworthiness metrics for social data - Adult Census.}}
    \label{fig:2dims_clean_rob_fairness}
    \vspace{-8mm}
\end{figure*}

\subsection{Accuracy and Robustness on Image 
 and Scientific Datasets}

\label{sec:robust_empirical}

We evaluate VTruST on three image datasets: CIFAR10 \citep{krizhevsky2009learning} , MNIST \citep{deng2012mnist} and Tinyimagenet \citep{le2015tiny} using ResNet-18 \citep{he2016deep}. For evaluation, we use the standard accuracy (SA) computed on the clean test sets and the robust accuracy (RA) computed on the corrupted test sets, CIFAR-10-C, Tiny ImageNet-C \citep{hendrycks2019robustness} and MNIST-C \citep{DBLP:journals/corr/abs-1906-02337}. While augmentation leads to robustness \citep{rebuffi2021data}, it also leads to a large dataset with redundancy. 
We use VTruST-R to select high-quality subset from augmented data.
Empirically we find that creation of augmented data by sampling images based on how difficult an augmentation is (\textit{Sampled Augmentation (SAug)}) leads to better performance compared to uniform selection across augmentations (\textit{Uniform Augmentation (UAug)}). We describe SAug (Algorithm \ref{algo:augs_detect}) and its comparison with UAug in Table \ref{tab:unifvssample} in the Appendix. Next, we define the baselines. 
(i) \textit{Clean-ST}: Unaugmented training dataset.
(ii) \textit{Uniform Augmentation (UAug)}
(iii) \textit{Sampled Augmentation (SAug)}
(iv) \textit{SSR} \citep{roh2021sample}: Training subset using the robustness objective function.
(v) \textit{AugMax} \citep{wang2021augmax}. 



\noindent \textbf{Robustness and Accuracy comparison (VTruST-R) with baselines:} We compare VTruST-R with the baselines in Table \ref{tab:robusteval} where it can be seen that model trained on clean datasets (\textit{Clean-ST}) performs abysmally in terms of RA, indicating the need of data augmentations.
VTruST-R is seen to outperform AugMax in most of the scenarios, thus indicating that data-centric approaches help in creating quality training datasets.

\begin{table*}[]
\centering
\tiny
%
\captionsetup{font=scriptsize}
\caption{\textbf{Comparison of VTruST-R over varying subset sizes for robustness evaluation. The numbers in brackets indicate the difference with the second best among baselines.}}

\label{tab:robusteval}
\begin{tabular}{|p{1.7cm}|lll|lll|lll|}
\hline
\multicolumn{1}{|c|}{\multirow{2}{*}{\textbf{Methods}}}        & \multicolumn{3}{c|}{\textbf{MNIST}}                                                                                                                                                                                                                         & \multicolumn{3}{c|}{\textbf{CIFAR10}}                                                                                                                                                                                                                       & \multicolumn{3}{c|}{\textbf{TinyImagenet}}                                                                                                                                                                                                                  \\ \cline{2-10} 
\multicolumn{1}{|c|}{}                                         & \multicolumn{1}{c|}{\textbf{\begin{tabular}[c]{@{}c@{}}\#Data\\ points\end{tabular}}} & \multicolumn{1}{c|}{\textbf{SA}} & \multicolumn{1}{c|}{\textbf{RA}} & \multicolumn{1}{c|}{\textbf{\begin{tabular}[c]{@{}c@{}}\#Data\\ points\end{tabular}}} & \multicolumn{1}{c|}{\textbf{SA}} & \multicolumn{1}{c|}{\textbf{RA}} & \multicolumn{1}{c|}{\textbf{\begin{tabular}[c]{@{}c@{}}\#Data\\ points\end{tabular}}} & \multicolumn{1}{c|}{\textbf{SA}} & \multicolumn{1}{c|}{\textbf{RA}} \\ \hline Clean-ST                                                         & \multicolumn{1}{l|}{60K}                                                                     & \multicolumn{1}{l|}{99.35}       & \multicolumn{1}{l|}{87.00}                                                                                             & \multicolumn{1}{l|}{50K}                                                                     & \multicolumn{1}{l|}{95.64}       & \multicolumn{1}{l|}{83.95}                                                                                             & \multicolumn{1}{l|}{100K}                                                                     & \multicolumn{1}{l|}{63.98}            & \multicolumn{1}{l|}{23.36}                                                                                                                                                          \\ \hline 
AugMax                                                         & \multicolumn{1}{l|}{240K}                                                                     & \multicolumn{1}{l|}{97.62}       & \multicolumn{1}{l|}{88.79}                                                                                             & \multicolumn{1}{l|}{200K}                                                                     & \multicolumn{1}{l|}{94.74}       & \multicolumn{1}{l|}{86.44}                                                                                             & \multicolumn{1}{l|}{400K}                                                                     & \multicolumn{1}{l|}{54.82}            & \multicolumn{1}{l|}{40.98}                                                                                                                       \\ \hline \hline
\rowcolor{Emerald} SAug & \multicolumn{1}{l|}{260K}                                                                     & \multicolumn{1}{l|}{\makecell{\textbf{99.36}\\(1.74)}}       & \multicolumn{1}{l|}{\makecell{\textbf{97.31}\\(8.52)}}                                                                                              & \multicolumn{1}{l|}{200K}                                                                     & \multicolumn{1}{l|}{\makecell{\textbf{94.9}\\(0.16)}}        & \multicolumn{1}{l|}{\makecell{\textbf{90.13}\\(3.69)}}                                                                                           & \multicolumn{1}{l|}{300K}                                                                     & \multicolumn{1}{l|}{\makecell{\textbf{62.04}\\(7.22)}}        & \multicolumn{1}{l|}{\makecell{\textbf{42.04}\\(1.06)}}       \\ \cline{1-10} \hline 
\multicolumn{10}{|c|} {\textbf{After subset selection from SAug}}  \\ \cline{1-10} \hline
SSR:40\%                                                       & \multicolumn{1}{l|}{104K}                                                                     & \multicolumn{1}{l|}{98.98}       & \multicolumn{1}{l|}{94.96}                                                                                              & \multicolumn{1}{l|}{80K}                                                                     & \multicolumn{1}{l|}{93.3}       & \multicolumn{1}{l|}{85.73}                                                                                           & \multicolumn{1}{l|}{120K}                                                                      & \multicolumn{1}{l|}{32.82}            & \multicolumn{1}{l|}{24.42}                                                    \\ \hline \rowcolor{LimeGreen}
VTruST-R:40\%                                                     & \multicolumn{1}{l|}{104K}                                                                     & \multicolumn{1}{l|}{\makecell{\textbf{99.04}\\(0.06)}}       & \multicolumn{1}{l|}{\makecell{\textbf{96.29}\\(1.33)}}                                                                                              & \multicolumn{1}{l|}{80K}                                                                     & \multicolumn{1}{l|}{\textbf{94.74}}      & \multicolumn{1}{l|}{\makecell{\textbf{88.23}\\(1.79)}}                                                                                            & \multicolumn{1}{l|}{120K}                                                                      & \multicolumn{1}{l|}{\makecell{\textbf{57.3}\\(2.48)}}            & \multicolumn{1}{l|}{39.69}                          
\\ \hline \hline
SSR:60\%                                                       & \multicolumn{1}{l|}{156K}                                                                     & \multicolumn{1}{l|}{99.07}       & \multicolumn{1}{l|}{96.53}                                                                                              & \multicolumn{1}{l|}{120K}                                                                     & \multicolumn{1}{l|}{93.77}       & \multicolumn{1}{l|}{88.0}                                                                                             & \multicolumn{1}{l|}{180K}                                                                      & \multicolumn{1}{l|}{41.94}            & \multicolumn{1}{l|}{30.07}                                                      \\ \hline \rowcolor{LimeGreen}
VTruST-R:60\%                                                     & \multicolumn{1}{l|}{156K}                                                                     & \multicolumn{1}{l|}{\makecell{\textbf{99.12}\\(0.05)}}       & \multicolumn{1}{l|}{\makecell{\textbf{97.09}\\(0.56)}}                                                                                              & \multicolumn{1}{l|}{120K}                                                                     & \multicolumn{1}{l|}{\makecell{\textbf{94.77}\\(0.03)}}       & \multicolumn{1}{l|}{\makecell{\textbf{89.21}\\(1.21)}}                                                                                        & \multicolumn{1}{l|}{180K}                                                                      & \multicolumn{1}{l|}{\makecell{\textbf{60.88}\\(6.03)}}            & \multicolumn{1}{l|}{\makecell{\textbf{41.50}\\(0.52)}}                                                                                             \\ \hline
\end{tabular}
\vspace{-4mm}
\end{table*}


\noindent \textbf{Scientific datasets} :
We analyzed the performance of VTruST-R on binary class scientific datasets - Spinodal and EOSL \citep{Benato_2022} that have 29,000 samples with 400 features and 180,000 samples with 576 features respectively. We used the experimental setup as \citep{Benato_2022} for evaluation. Table \ref{tab:scientific-ds} shows that VTruST-R (using label flipping for robustness) performs close to the wholedata in standard accuracy (SA) and better in terms of robust accuracy (RA). The remaining results can be found in the Appendix.



\begin{table}[]
\tiny
\captionsetup{font=scriptsize}
\caption{\textbf{Performance comparison on scientific datasets}}
\label{tab:scientific-ds}
\begin{tabular}{|l|lllllll|llll|}
\hline
\multirow{2}{*}{\textbf{Metrics}} & \multicolumn{7}{c|}{\textbf{Spinodal}}                                                                                                                                                                                                                                                                                                                                                                                                                                                                                       & \multicolumn{4}{c|}{\textbf{EOSL}}                                                                                                                                                                                                                                                                                                                                                                                                                                                                                         \\ \cline{2-12} 
                                  & \multicolumn{1}{l|}{\begin{tabular}[c]{@{}l@{}}Whole\\ data\end{tabular}} & \multicolumn{1}{l|}{\begin{tabular}[c]{@{}l@{}}Rand\\ 40\%\end{tabular}} & \multicolumn{1}{l|}{\begin{tabular}[c]{@{}l@{}}SSFR\\ 40\%\end{tabular}} & \multicolumn{1}{l|}{\begin{tabular}[c]{@{}l@{}}\cellcolor{LimeGreen}VTruST\\\cellcolor{LimeGreen}-R 40\%\end{tabular}} & \multicolumn{1}{l|}{\begin{tabular}[c]{@{}l@{}}Random\\ 60\%\end{tabular}} & \multicolumn{1}{l|}{\begin{tabular}[c]{@{}l@{}}SSFR\\ 60\%\end{tabular}} & \begin{tabular}[c]{@{}l@{}} \cellcolor{LimeGreen}VTruST\\ \cellcolor{LimeGreen}-R 60\%\end{tabular} & \multicolumn{1}{l|}{\begin{tabular}[c]{@{}l@{}}Whole\\ data\end{tabular}} & \multicolumn{1}{l|}{\begin{tabular}[c]{@{}l@{}}Rand\\ 40\%\end{tabular}} & \multicolumn{1}{l|}{\begin{tabular}[c]{@{}l@{}}SSFR\\ 40\%\end{tabular}} & \multicolumn{1}{l|}{\begin{tabular}[c]{@{}l@{}}\cellcolor{LimeGreen}VTruST\\ \cellcolor{LimeGreen}-R 40\%\end{tabular}} \\ \hline
SA                                & \multicolumn{1}{l|}{\textbf{83.08}}                                       & \multicolumn{1}{l|}{73.05}                                               & \multicolumn{1}{l|}{74.84}                                               & \multicolumn{1}{l|}{\textbf{\cellcolor{LimeGreen}80.33}}                                                 & \multicolumn{1}{l|}{77.06}                                                 & \multicolumn{1}{l|}{78.94}                                               & \textbf{\cellcolor{LimeGreen}81.93}                                                 & \multicolumn{1}{l|}{70.01}                                                & \multicolumn{1}{l|}{63.74}                                                    & \multicolumn{1}{l|}{62.40}                                                    & \multicolumn{1}{l|}{\textbf{\cellcolor{LimeGreen}66.10}}                                                                                                       \\ \hline
RA                                & \multicolumn{1}{l|}{76.89}                                                & \multicolumn{1}{l|}{61.11}                                               & \multicolumn{1}{l|}{62.32}                                               & \multicolumn{1}{l|}{\cellcolor{LimeGreen}\textbf{78.36}}                                                 & \multicolumn{1}{l|}{75.11}                                                 & \multicolumn{1}{l|}{75.18}                                               & \cellcolor{LimeGreen}\textbf{80.41}                                                 & \multicolumn{1}{l|}{66.72}                                                & \multicolumn{1}{l|}{60.04}                                                    & \multicolumn{1}{l|}{56.90}                                                    & \multicolumn{1}{l|}{\textbf{\cellcolor{LimeGreen}65.27}}                                                                                                    \\ \hline
\end{tabular}
\vspace{-8mm}
\end{table}






\vspace{-4mm}
\subsection{Data-centric analysis: Post hoc explanation}

In this section, we explore the characteristics of the selected samples to justify their quality.

\noindent
\textbf{Explanation for fairness:}
We use the metric \textit{Counterfactual Token Fairness Gap (CF-Gap)}\citep{garg2019counterfactual} 
for our evaluation. Given a selected instance $x$, we generate a counterfactual instance $x'$ by altering its sensitive attribute and define CF-Gap$(x)$ as $\|f(x_i)-f(x_i')|$ where $f(x)$ corresponds to the model confidence on the target label.
We plot the distribution of CF-Gap in Figure \ref{fig:indiv_cfgap_adult}. It can be observed that VTruST-F acquires the least value, justifying its retainment of fair subsets leading to fair models.
We show 10 anecdotal samples from the Adult Census dataset in Table \ref{tab:anec_cfgap} on the basis of high \textit{CF-Gap} 
and we can observe that SSFR has a large number of redundant samples with similar attribute values (highlighted)
while VTruST-F which anyway has relatively lower CF-gap contains a diverse set of samples. 

\begin{minipage}{0.25\linewidth}
    \captionsetup{font=scriptsize}
    \captionof{figure}{\textbf{Box plot representation of CF-gap.}}
    \label{fig:indiv_cfgap_adult}
    \begin{subfigure}{\columnwidth}
     \includegraphics[width=1.2\textwidth,height=3cm]{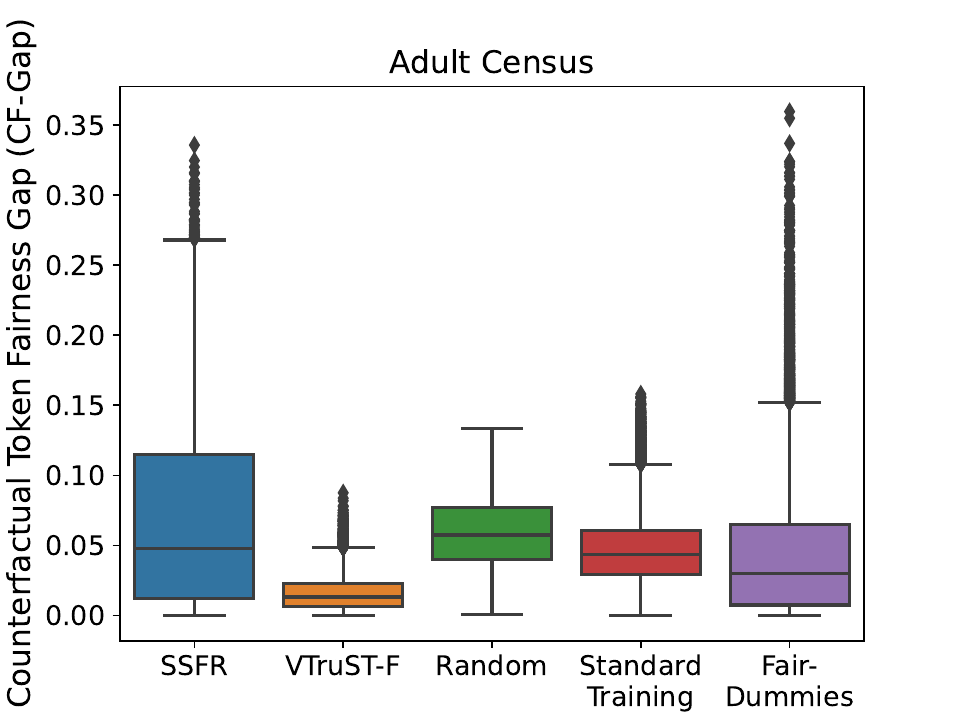}
    
    \end{subfigure}
    
\end{minipage}
\hspace{2mm}
\begin{minipage}{0.6\linewidth}
\centering
\tiny
\captionsetup{font=scriptsize}
\captionof{table}{\textbf{Sample instances with High Counterfactual Token Fairness Gap}}
\label{tab:anec_cfgap}
\begin{minipage}{0.52\linewidth}
\begin{tabular}{|c|| c | c | c | c |} \hline
\multicolumn{5}{|c|}{\textbf{VTruST-F}}  \\ \cline{1-5}
 \textit{\textbf{Feat}} & \textbf{Rel} & \textbf{Race} & \textbf{Sex} & \textbf{NC} \\ 
 \hline\hline
\textbf{$D_1$}  & ORel & B & F & JM \\ 
\textbf{$D_2$}   & NIF & W & M & US \\  
\textbf{$D_3$} & NIF & W & M & US \\ 
\textbf{$D_4$} & OC & API & F & TW \\ 
\textbf{$D_5$}  & Husb & W & M & US \\ 
\textbf{$D_6$}  & UnM & W & F & US \\ 
\textbf{$D_7$}  & Wife & W & F & US \\ 
\textbf{$D_8$}  & OC & W & M & US \\ 
\textbf{$D_9$}  & NIF & AIE & F & Col \\ 
\textbf{$D_{10}$}  & UnM & W & F & DE \\ 

\hline
\end{tabular}
\end{minipage}%
\begin{minipage}{0.49\linewidth}
\centering
\begin{tabular}{|c|| c | c | c | c |} 
\hline
\multicolumn{5}{|c|}{\textbf{SSFR}}  \\ \cline{1-5} 
 \textit{\textbf{Feat}} & \textbf{Rel} & \textbf{Race} & \textbf{Sex} & \textbf{NC} \\ 
 \hline\hline
 \textbf{$D_1$}  & \cellcolor{blue!25}Husb & \cellcolor{blue!25}W & \cellcolor{blue!25}M & \cellcolor{blue!25} US \\ 
 \textbf{$D_2$}  & \cellcolor{blue!25}Husb & \cellcolor{blue!25}W & \cellcolor{blue!25}M & \cellcolor{blue!25}US \\ 
\textbf{$D_3$}  & \cellcolor{blue!25}Husb & \cellcolor{blue!25}W & \cellcolor{blue!25}M & \cellcolor{blue!25}US \\ 
\textbf{$D_4$}  & \cellcolor{blue!25}Husb & \cellcolor{blue!25}W & \cellcolor{blue!25}M & \cellcolor{blue!25}US \\ 
\textbf{$D_5$}  & Husb & W & M & US \\ 
\textbf{$D_6$}  & \cellcolor{blue!25}Husb & \cellcolor{blue!25}W & \cellcolor{blue!25}M & \cellcolor{blue!25}US \\ 
\textbf{$D_7$} & NIF & W & M & US \\
\textbf{$D_8$} & OC & W & M & US \\
\textbf{$D_9$}  & \cellcolor{blue!25}Husb & \cellcolor{blue!25}W & \cellcolor{blue!25}M & DE \\ 
\textbf{$D_{10}$} & OC & W & M & US \\ 
\hline
\end{tabular}
\end{minipage}
\end{minipage}

\textbf{Explanation for robustness:} Delving into the literature \citep{swayamdipta2020dataset,huang2018cost}, we pick two measures - \textit{uncertainty} and \textit{distinctiveness}. Having a set of hard-to-learn and distinguishable samples in the subsets makes the model more generalizable and robust. 
We quantify uncertainty of an instance $x$ as predictive entropy ($-f(x)logf(x)$) and distinctiveness as $\mathbb{E}_{e \in \cS} dist(fv(x),fv(e))$ where $dist(,)$ is the euclidean distance and $fv(.)$ is the feature from the model's penultimate layer.
Based on the data maps in Figure \ref{fig:datamap} in Appendix, we show anecdotal samples in Figure \ref{fig:anec_high_low_tim} having High Distinctiveness-High Uncertainty (HD-HU). The anecdotal samples and the histogram visualization show that VTruST-R selects diverse samples with difficult augmentations like impulse noise and glass blur, while similar(mostly white-background) and no-noise or easier augmentation-based samples like brightness are more observed in SSR samples, thus justifying the robust selection across augmentations using VTruST-R.

\begin{figure}[h!]
    \begin{subfigure}{\textwidth}
    \includegraphics[width=0.6\linewidth,height=2cm]{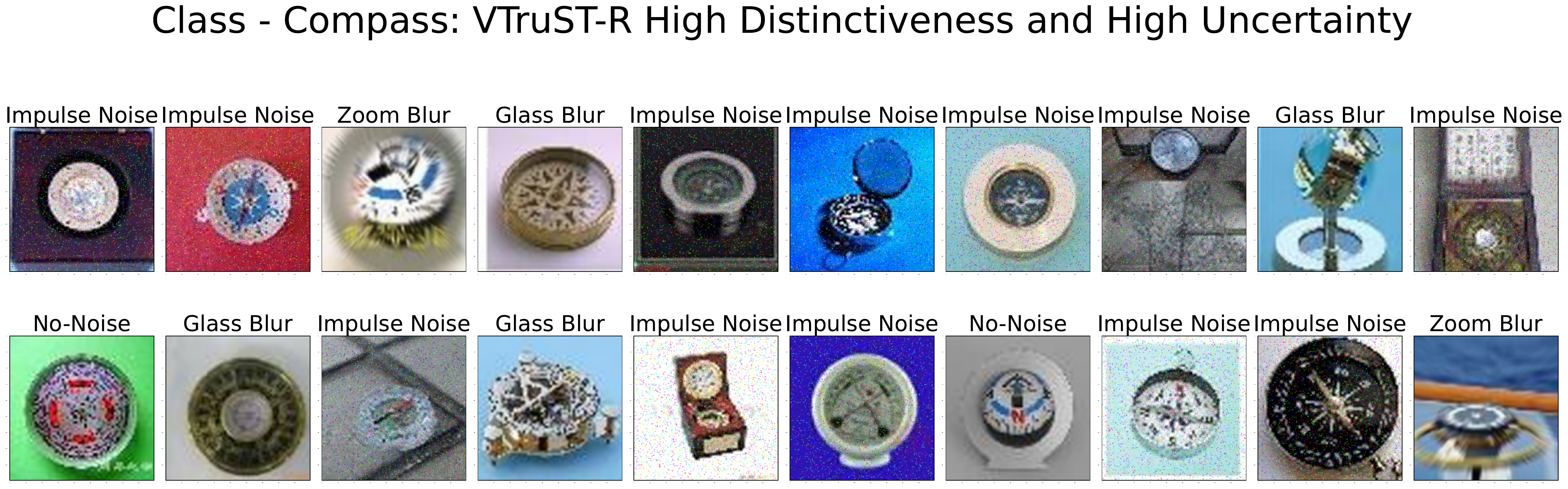}
    \end{subfigure}
    \begin{subfigure}{\textwidth}
    \includegraphics[width=0.35\linewidth,height=2cm]{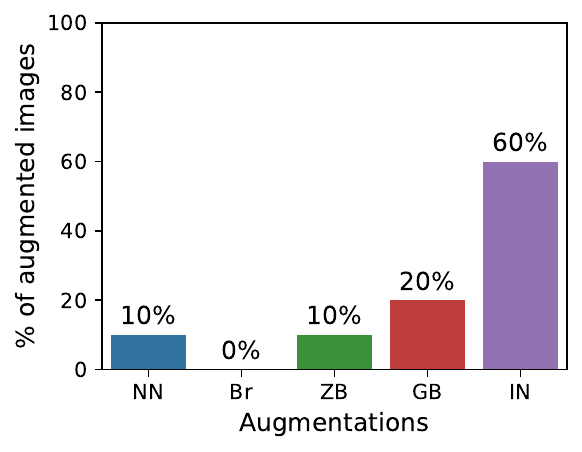}
    \end{subfigure} 

    \hspace{10mm}

    \begin{subfigure}{\textwidth}
    \includegraphics[width=0.6\linewidth,height=2cm]{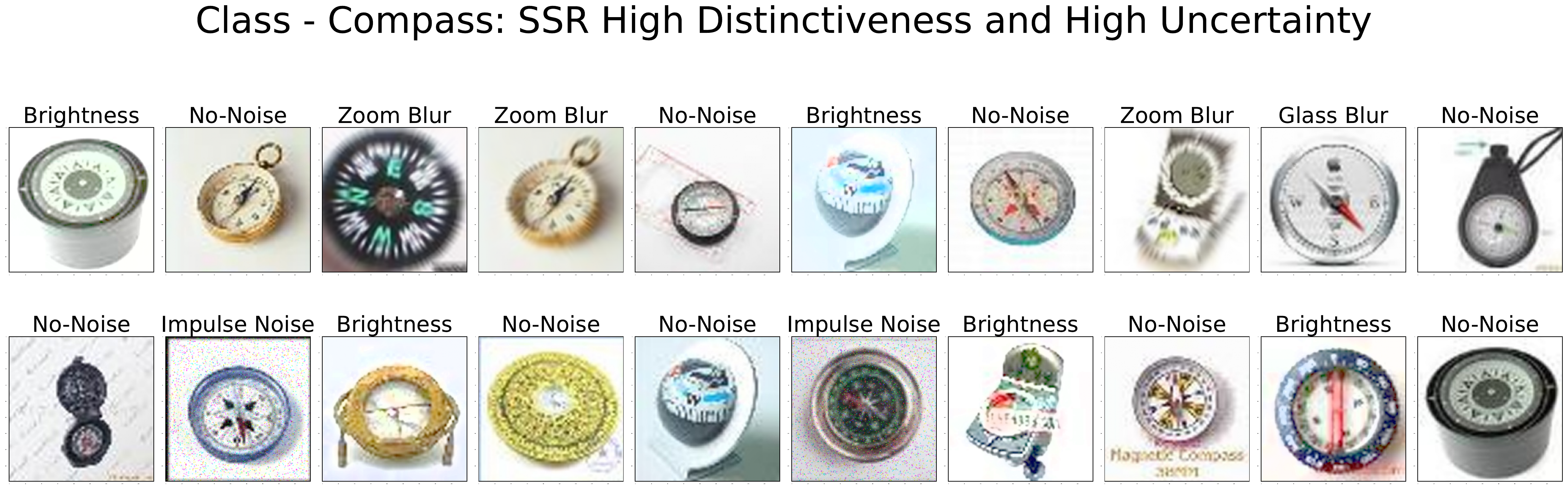}
    \end{subfigure}
    \begin{subfigure}{\textwidth}
    \includegraphics[width=0.35\linewidth,height=2cm]{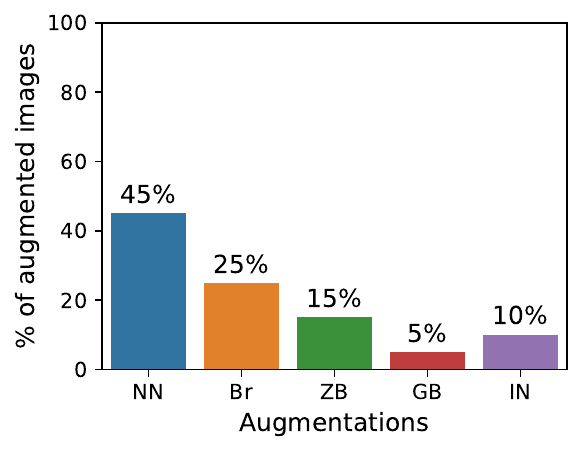}
    \end{subfigure}
    \captionsetup{font=scriptsize}
    \caption{\textbf{Anecdotal samples from VTruST-R \& SSR with High Distinctiveness and Uncertainty from TinyImagenet for class Compass.}}
    \label{fig:anec_high_low_tim}
    \vspace{-8mm}
\end{figure}

\section{Discussion and Related works}

Existing works on trustworthy AI \citep{liang2022advances} have focussed on designing fair \citep{pmlr-v28-zemel13,romano2020achieving,sattigeri2022fair,chuang2021fair} and robust \citep{wang2021augmax,Chen_2022_CVPR} models. Several works have also been done to inevstigate the tradeoffs between pairs of trustworthiness metrics - fairness vs accuracy \citep{roh2020fairbatch}; robustness vs accuracy \citep{pmlr-v162-pang22a,hu2023understanding} ; fairness vs robustness \citep{roh2021sample}. The closest approach to our method is that of \citep{roh2021sample} which selects fair and robust samples by enforcing a constraint on the number of selected samples per class. However, none of the above methods have the flexibility of a user-controllable tradeoff between trustworthiness metrics. Besides, they impose an additional constraint on the training objective that may lead to a potential bias.  Hence, there arises a need for a paradigm shift from model-centric to data centric approaches that would look at the input space and sample the quality datapoints with potentially less bias.

The existing works in data-centric AI (DCAI) have explored data valuation approaches for obtaining quality data. \citep{swayamdipta2020dataset,ethayarajh2022understanding, NEURIPS2022_95b6e2ff,seedat2022data} work on data quality measures 
to determine hard and easy samples. The other category of valuation methods are mostly based on Shapley values \citep{ghorbani2019data,wang2022data}, influence functions \citep{koh2017understanding,park2023trak}, reinforcement learning \citep{yoon2020data}, gradient-based approximations \citep{NEURIPS2020_61d77652,paul2021deep,killamsetty2021grad,das2021finding} and training free scores \citep{just2023lava,nohyun2022data,wu2022davinz}. However, all the methods only account for accuracy and none of the other trustworthiness metrics. Our proposed method, VTruST, lies in an intersectional area between trustworthy AI and data valuation. To the best of our knowledge, ours is one of the first works in DCAI that develops a controllable framework to provide a tradeoff across different trustworthiness metrics (fairness, robustness and accuracy) leading to desired subsets in an online training paradigm.

\newpage
\section*{Reproducibility Statement}

We run all our experiments on publicly available datasets and thus all our results can be seamlessly reproduced. 
The code is available at \url{https://github.com/dmlr-vtrust/VTruST/}.
Details on model architectures and datasets are provided in the main paper. The remaining details for obtaining reproducible results can be found in the Appendix.

\vskip 0.2in
\bibliography{sample}

\newpage
\appendix
\section{Appendix}

\section*{1. Empirical evaluation on Social Data}

\begin{figure}[h!]

    \centering
\begin{subfigure}{\columnwidth}
    \centering    \includegraphics[width=0.48\textwidth,height=3cm]{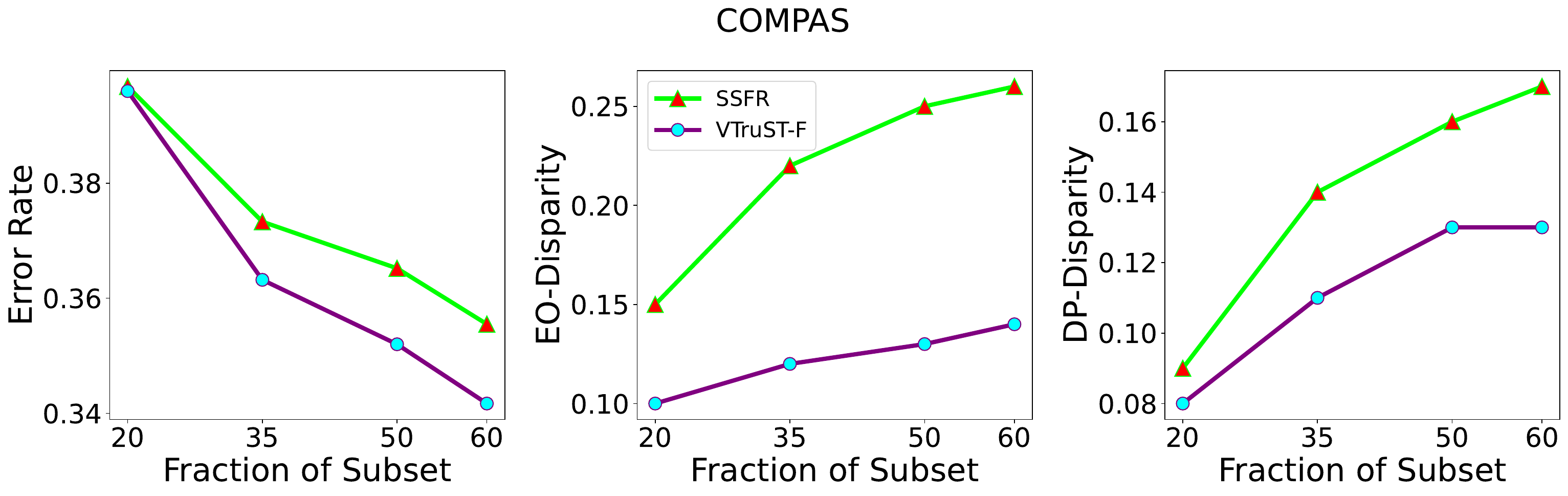}
    \end{subfigure}
    \begin{subfigure}{\columnwidth} 
    \centering
\includegraphics[width=0.48\textwidth,height=3cm]{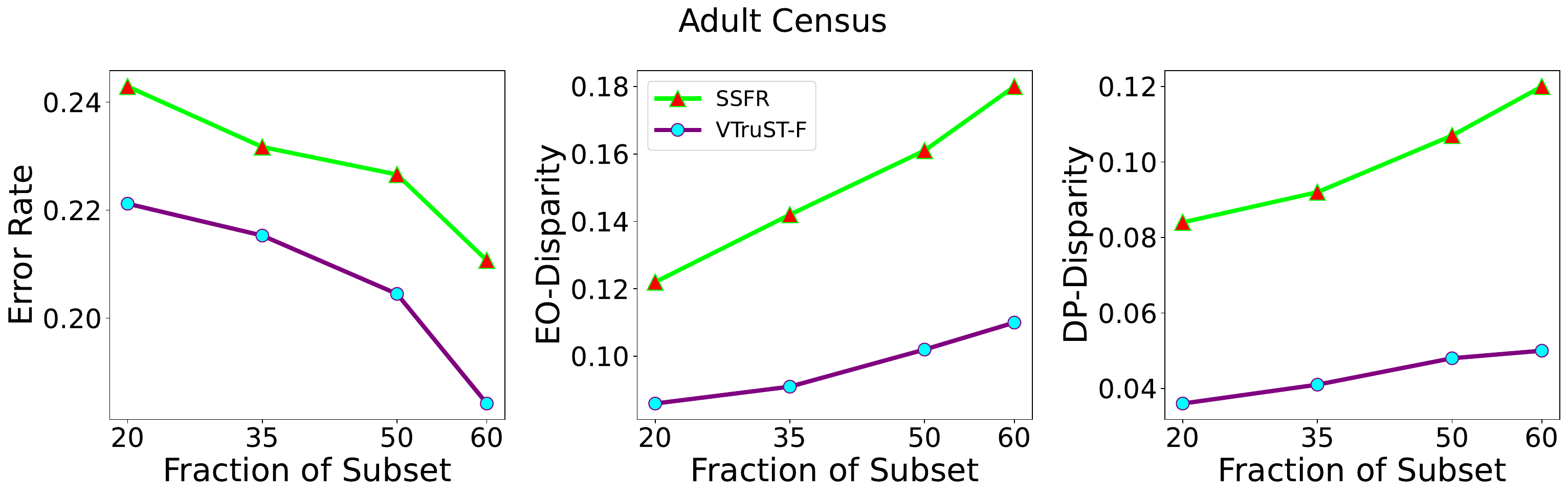}
    \end{subfigure}
    
    \begin{subfigure}{\columnwidth}
    \centering    \includegraphics[width=0.48\textwidth,height=3.5cm]{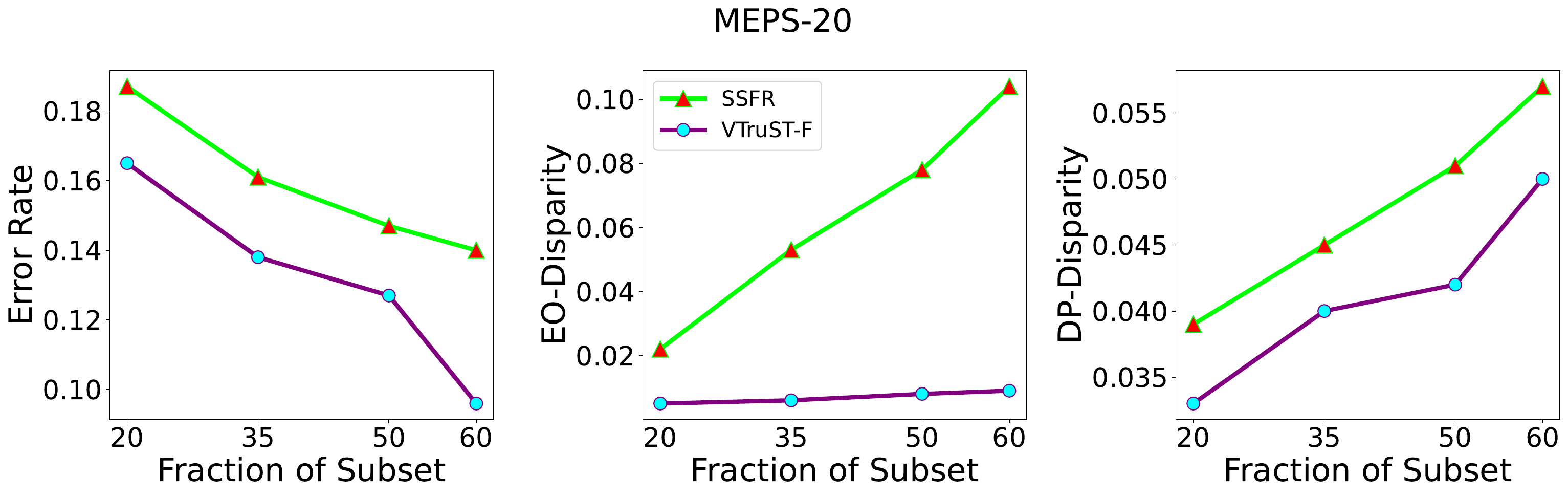}
    \end{subfigure}
    
    \caption{\textbf{Varying fraction of subsets: We report the ER and disparities for different subset sizes selected by the proposed method VTruST-F and SSFR. It can be observed that the proposed method always stays below the baselines in terms of error rate and disparity measures.}}
    \label{fig:varying_frac}
\end{figure}

\begin{figure}[h!]
  \centering
    \begin{subfigure}{\textwidth}
    \centering
    \includegraphics[width=0.48\textwidth,height=3.5cm]{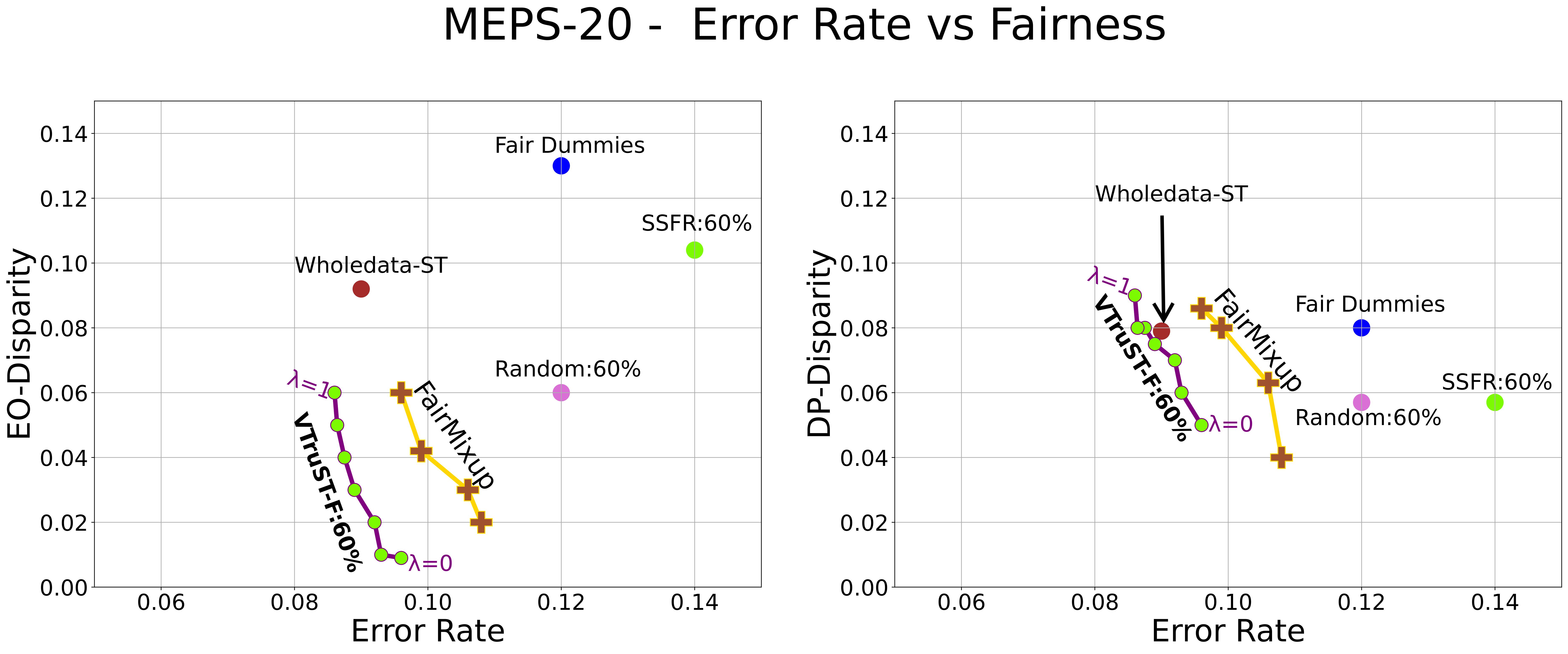}
    \end{subfigure}
    \begin{subfigure}{\textwidth}
    \centering    \includegraphics[width=0.48\textwidth,height=3.5cm]{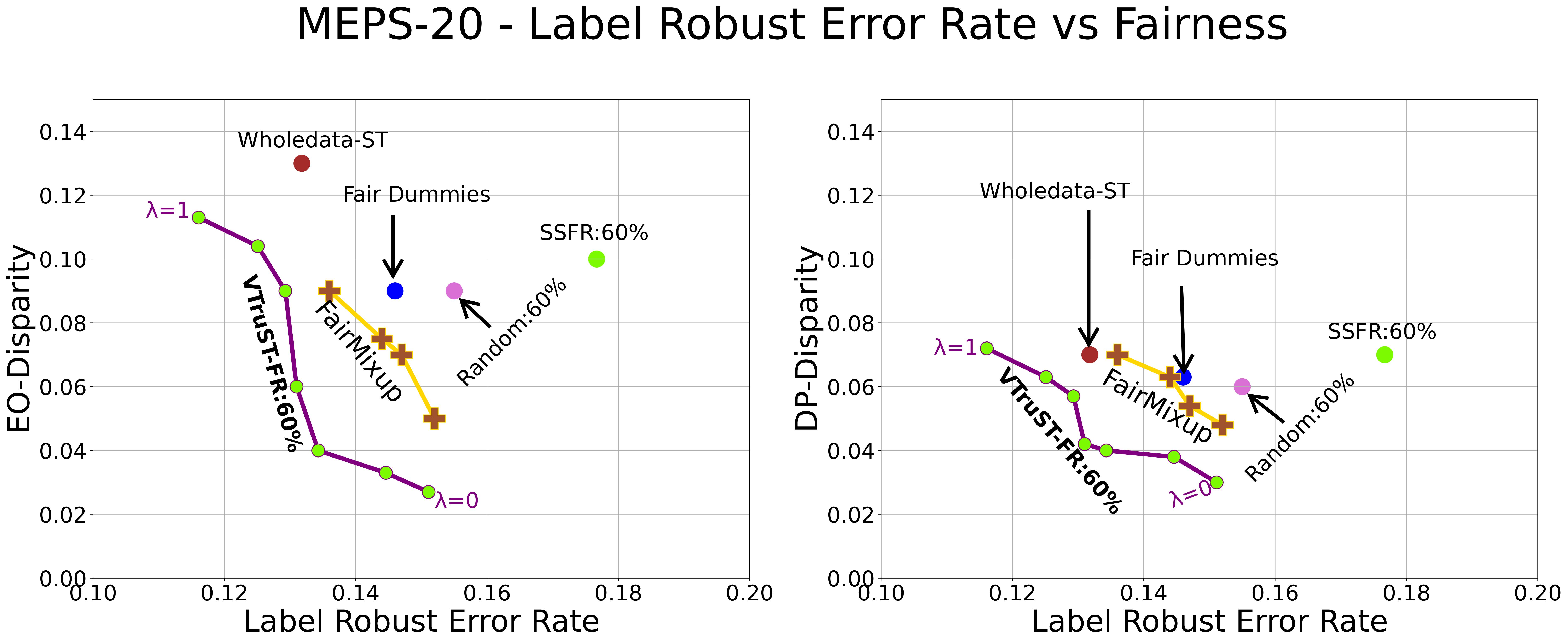}
    \end{subfigure}
  

    \begin{subfigure}{\textwidth}
    \centering
    \includegraphics[width=0.48\textwidth,height=3.5cm]{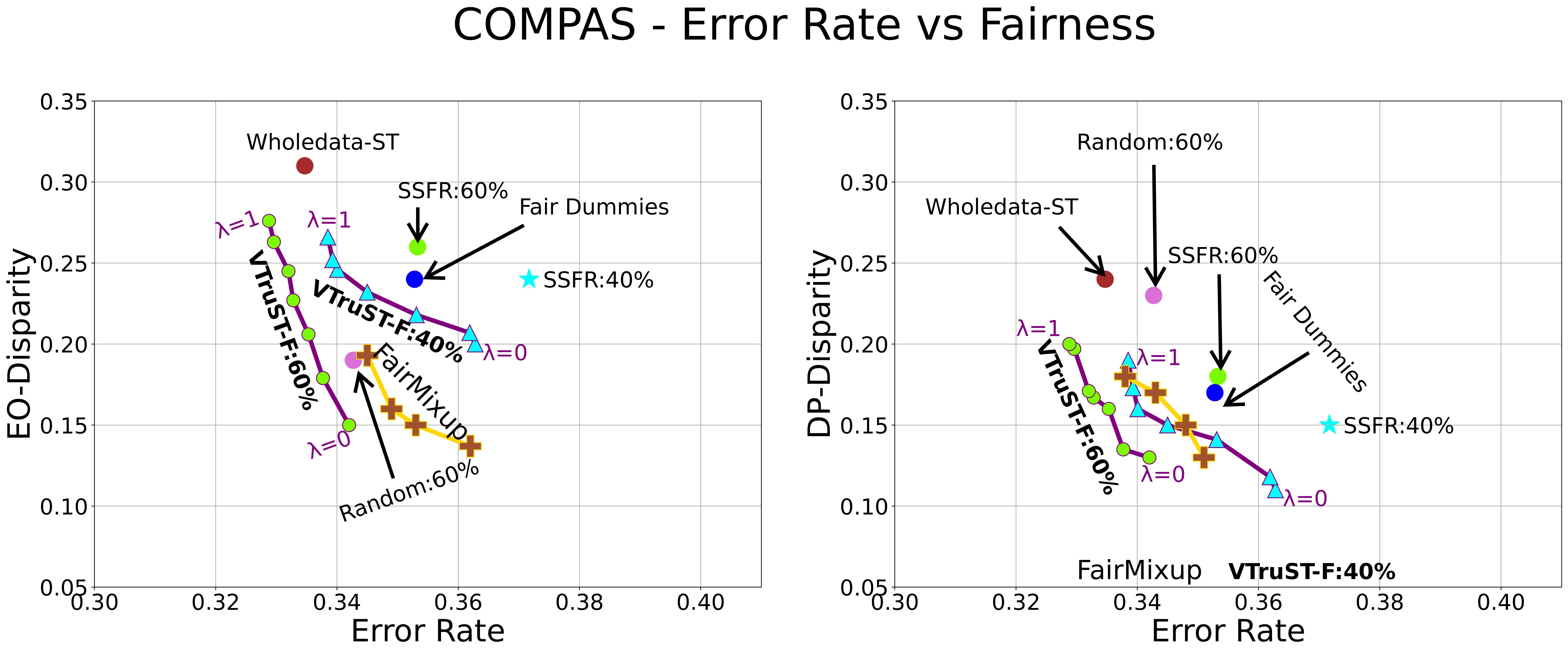}
    \end{subfigure}
    \begin{subfigure}{\textwidth}
    \centering    \includegraphics[width=0.48\textwidth,height=3.5cm]{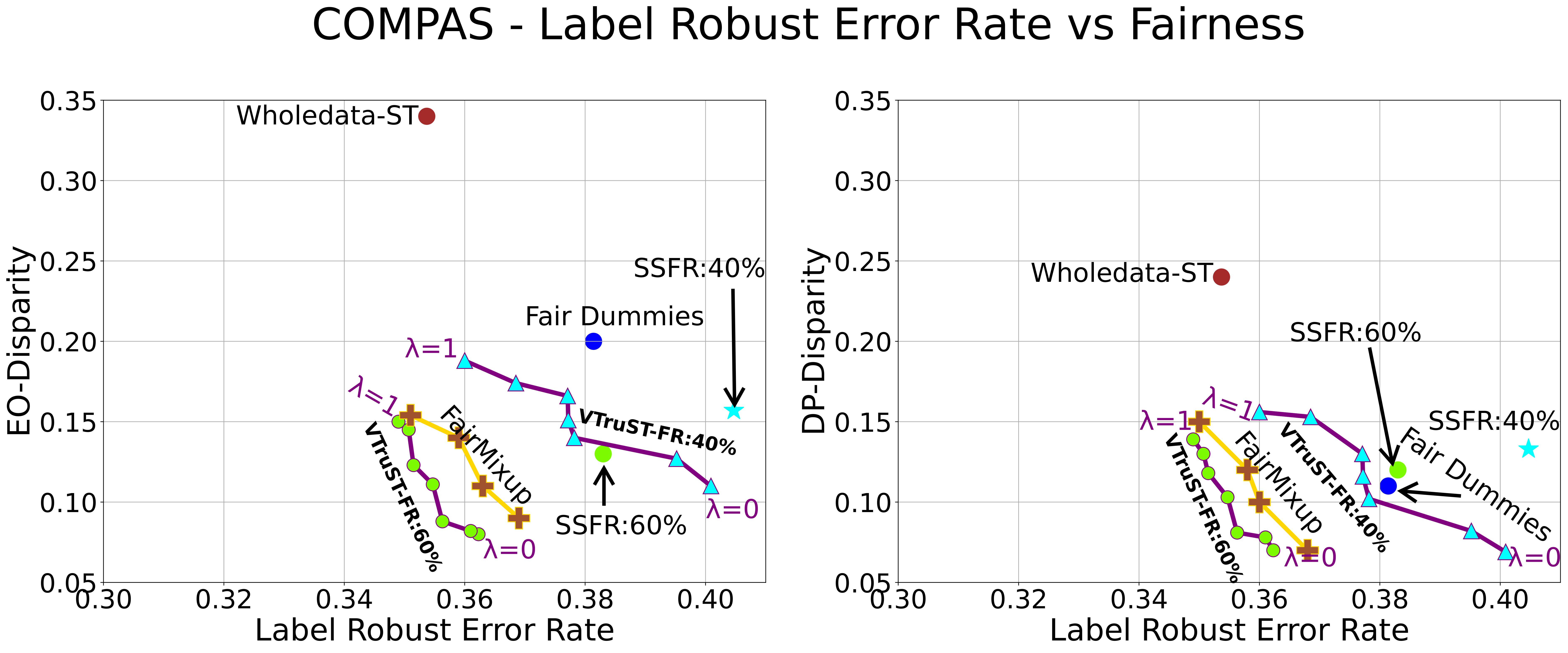}
    \end{subfigure}

    \caption{\textbf{Error Rate-Fairness and Robustness-Fairness tradeoff in clean and augmented data setup : We show the performance of the methods w.r.t the two dimensions - \textit{Performance and Disparity} and can observe that the proposed method VTruST lies relatively on the \textcolor{blue}{bottom left region} (\textit{low error rate or robust error rate-low disparity}) with disparity being the lowest for $\lambda=0$. Higher weightage to $\lambda$ leads to a low error rate or robust error rate for the same fraction and increasing disparity.}}
    \label{fig:2dims_fairness_remain}
\end{figure}

In this section, we report the empirical results for the fairness value function. We show the variation in performance measures with varying subset fractions in Figure \ref{fig:varying_frac}. It can be clearly observed that VTruST-F outperforms SSFR and that VTruST-F has the lowest Error Rate (ER) and disparity measures across all the considered fractions. We show the pareto-frontal curve for both clean and noisy datasets from MEPS20 and COMPAS in Figure \ref{fig:2dims_fairness_remain} where we can observe that VTruST-F has the lowest disparity for $\lambda=0$ and lowest error rate for $\lambda=1$. It lies relatively in the bottom-left region compared to other baselines.

\section*{2. Sampled Augmentation - SAug}

\begin{figure}
    \centering
    \captionsetup{font=scriptsize}
    \captionof{figure}{\textbf{Performance of self-trained augmentation models on augmented test sets.}}
    \label{fig:heatmap}
    \begin{subfigure}{\textwidth}
     \hspace*{-1cm}
     \includegraphics[width=0.8\textwidth,height=5cm]{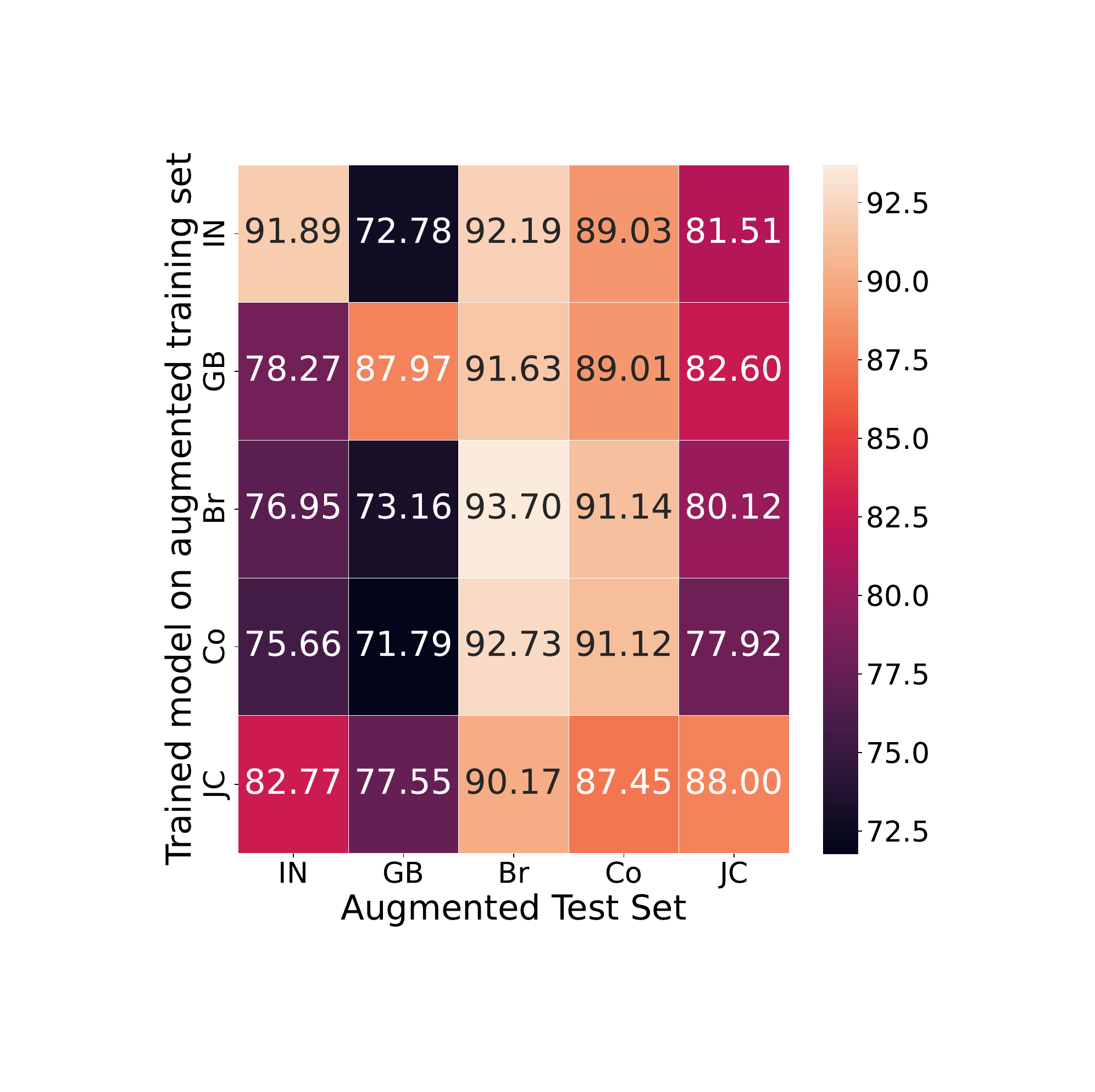}
    
    \end{subfigure}
\end{figure}    

\begin{algorithm}
\scriptsize
\caption{: \textbf{Sampling augmentations}}
\label{algo:augs_detect}
 \begin{algorithmic}[1]
 
 \STATE{\textbf{Input}:} $Mat_{M\times |A|}$ // Matrix[i,j] = Robust accuracy on test set augmented with corruption $j \in A$ using model $m \in M$ trained with data augmented with augmentation $i \in A$. $M = |A|+1$ where we also test using model trained on clean data. ; Iteration: $t=0$ ; Clean dataset: $D^t$ ;  \\
 \STATE $SN_j^t =  \frac{Mat[j,j]- \frac{\sum_{i \neq j}Mat[i,j]}{M-1}}{Mat[j,j]}$
 \STATE Normalise $SN_j^t \forall j \in A$ and sample that fraction of images for the respective augmentations from all classes uniformly and form $D^{t+1}$.
 \STATE Train on $D^{t+1}$ and obtain model $f$.
 \STATE Compute robust accuracy for each $j$ using trained model $f$.
  \STATE $SN_j^{t+1} =  Mat[j,j]- RA_j^{t+1}$
  \STATE $t = t+1$
  \STATE Repeat from line 3 till $\frac{\sum_j RA_j^t}{|A|}-\frac{\sum_j RA_j^{t-1}}{|A|} < \epsilon$
  \STATE{\textbf{Output}:} $D^t$
 \end{algorithmic}
\end{algorithm}

\begin{table}[]
\centering
\tiny
\captionsetup{font=scriptsize}
\captionof{table}{\textbf{Comparison of Uniform Augmentation with Sampled Augmentation.}}
\label{tab:unifvssample}
\tiny
\begin{tabular}{|l|l|l|l|l|l|l|}
\hline
\textbf{Metrics}                                                     & \begin{tabular}[c]{@{}l@{}}UAug (260K)\\ MNIST\end{tabular} & \begin{tabular}[c]{@{}l@{}}SAug (260K)\\ MNIST\end{tabular} & \begin{tabular}[c]{@{}l@{}}UAug (200K)\\ CIFAR10\end{tabular} & \begin{tabular}[c]{@{}l@{}}SAug (200K)\\ CIFAR10\end{tabular} & \begin{tabular}[c]{@{}l@{}}UAug (300K)\\ TinyImagenet\end{tabular} & \begin{tabular}[c]{@{}l@{}}SAug (300K)\\ TinyImagenet\end{tabular} \\ \hline
\textbf{\begin{tabular}[c]{@{}l@{}}Standard\\ Accuracy\end{tabular}} & 99.34                                                       & \begin{tabular}[c]{@{}l@{}}99.37\\ (0.14)\end{tabular}      & 94.84                                                         & \begin{tabular}[c]{@{}l@{}}94.9\\ (0.06)\end{tabular}         & 60.92                                                              & \begin{tabular}[c]{@{}l@{}}62.04\\ (1.12)\end{tabular}             \\ \hline
\textbf{\begin{tabular}[c]{@{}l@{}}Robust\\ Accuracy\end{tabular}}   & 97.12                                                       & \begin{tabular}[c]{@{}l@{}}97.31\\ (0.19)\end{tabular}      & 89.06                                                         & \begin{tabular}[c]{@{}l@{}}90.13\\ (1.07)\end{tabular}        & 26.87                                                              & \begin{tabular}[c]{@{}l@{}}42.04\\ (15.87)\end{tabular}            \\ \hline
\end{tabular}
\end{table}

\begin{figure}[h!]
    \centering
    \begin{subfigure}{\textwidth}
    \centering   \includegraphics[width=0.2\textwidth,height=3cm]{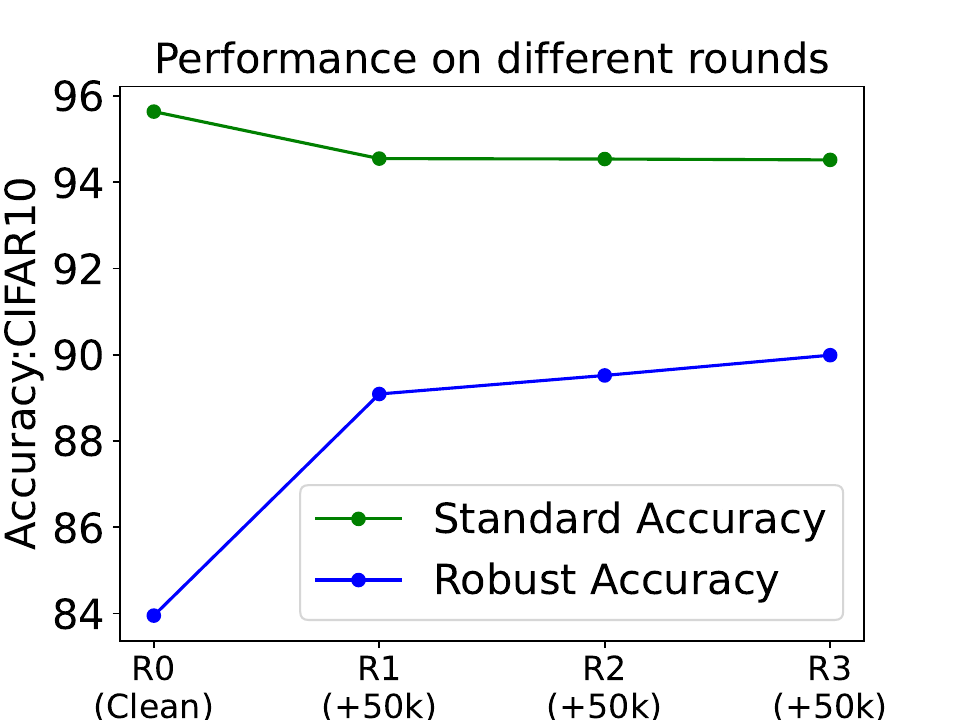}
    \end{subfigure}    
    \begin{subfigure}{\textwidth}
    \centering
    \includegraphics[width=0.2\textwidth,height=3cm]{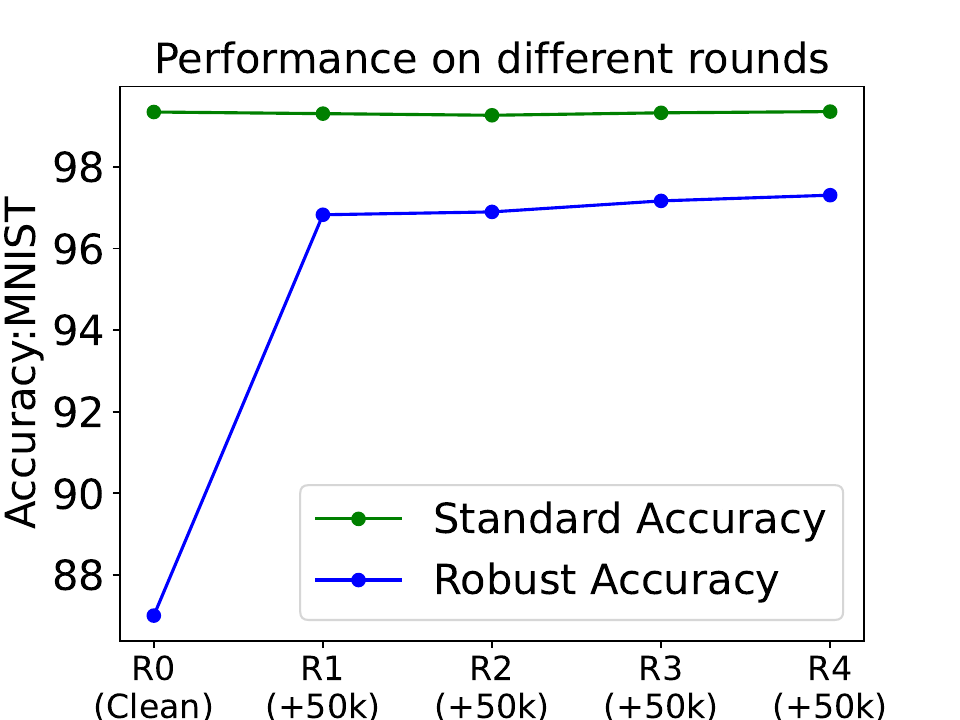}
    \end{subfigure}
    \begin{subfigure}{\textwidth}
    \centering
    \includegraphics[width=0.2\textwidth,height=3cm]{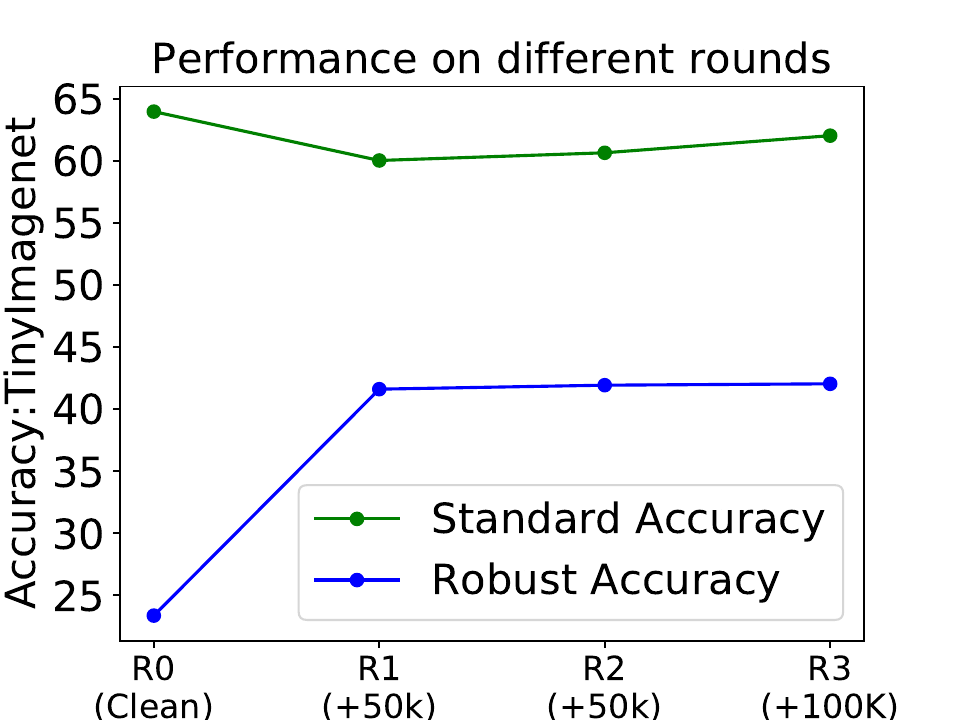}
    \end{subfigure} 
     \caption{\textbf{Mean accuracies across all augmentations over each round. R0 represents the accuracy obtained using model trained on clean data. The subsequent rounds are using augmented images obtained from our proposed sampling algorithm. The drop in Standard Accuracy is marginal, while the increase in Robust Accuracy is significant.}}
    \label{fig:aug_heatmap_rounds}
\end{figure}

Firstly, we look at the performance of the models across different augmentations that worked as an intuition for the sampling algorithm. Figure \ref{fig:heatmap} depicts the difference in performance across difference augmentations. The cells $(i,j)$ corresponds to performance of a model trained on augmentation $i$ and tested on augmentation $j$. The diagonal element correspond to the self trained augmentation accuracies that turn out to be the best for any augmentation.
Based on the intuition developed from the above heatmap, we present the pseudocode of our sampling augmentation in Algorithm \ref{algo:augs_detect}. We define Sampling Number (\textit{SN}) for augmentation $j$ as a normalized difference between the average RA for aug $j$ ($RA_j$) and the self-trained accuracy. We execute the algorithm and plot the standard and robust accuracies across the different rounds in Figure \ref{fig:aug_heatmap_rounds}. $R0$ corresponds to the round when we use the model trained on clean/non-augmented data. We can observe in Figure \ref{fig:aug_heatmap_rounds} that the similar pattern is observed across all the datasets where the standard accuracy gets compromised marginally with a significant improvement in robust accuracy, thus justifying the use of augmentations for robustness.

\section*{3. Empirical evaluation on scientific data}

We report the results on EOSL dataset for the 60\% subset in Table \ref{tab:scientific-ds-eosl}. It can be observed that VTruST-R performs better than the other baselines.

\begin{table}[]
\centering
\captionsetup{font=scriptsize}
\caption{\textbf{Performance comparison on scientific datasets}}
\label{tab:scientific-ds-eosl}
\begin{tabular}{|l|llll|}
\hline
\multirow{2}{*}{\textbf{Metrics}}                                                                                                                        & \multicolumn{4}{c|}{\textbf{EOSL}}                                                                                                                                                                                                                                                                                                                                                                                                                                                                                         \\ \cline{2-5} 
& \multicolumn{1}{l|}{\begin{tabular}[c]{@{}l@{}}Whole\\ data\end{tabular}} & \multicolumn{1}{l|}{\begin{tabular}[c]{@{}l@{}}Rand\\ 60\%\end{tabular}} & \multicolumn{1}{l|}{\begin{tabular}[c]{@{}l@{}}SSFR\\ 60\%\end{tabular}} & \begin{tabular}[c]{@{}l@{}}\cellcolor{LimeGreen}VTruST\\\cellcolor{LimeGreen}-R 60\%\end{tabular} \\ \hline
SA                                                                                 & \multicolumn{1}{l|}{70.01}                                                                                                   & \multicolumn{1}{l|}{64.94}                                               & \multicolumn{1}{l|}{64.64}                                               & \textbf{\cellcolor{LimeGreen}68.86}                                                 \\ \hline
RA                                                                                                              & \multicolumn{1}{l|}{66.72}                                                                                                    & \multicolumn{1}{l|}{61.55}                                               & \multicolumn{1}{l|}{62.43}                                               & \cellcolor{LimeGreen}\textbf{67.67}                                                 \\ \hline
\end{tabular}
\end{table}

\section*{4. Data centric explanation for Fairness}

\begin{figure}[h!]
    \centering
    \begin{subfigure}{\columnwidth}
    \centering        \includegraphics[width=0.31\textwidth,height=4cm]{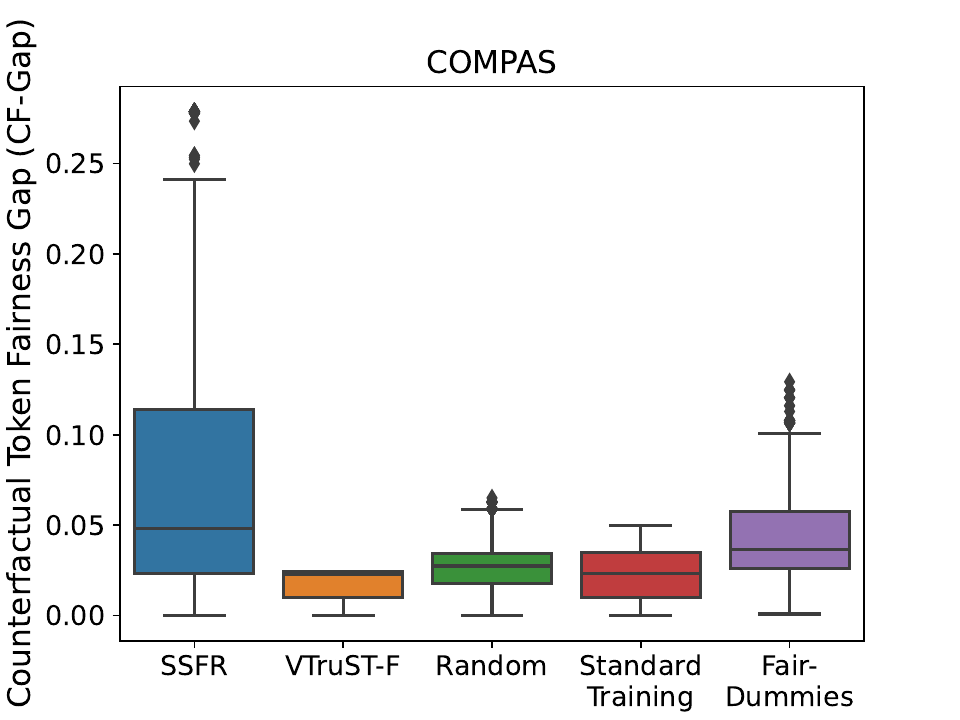}
    \end{subfigure}
    \caption{\textbf{Box plot representation of counterfactual token fairness gap on the selected subsets from VTruST-F and other baselines for COMPAS dataset.}}
    \label{fig:indiv_cfgap_predsens_syn}
\end{figure}

We report the CF-Gap for the COMPAS dataset in Figure \ref{fig:indiv_cfgap_predsens_syn} and show that the proposed algorithm has the lowest values of the measures (the lower, the fairer) compared to all other baselines.

\section*{5. Data centric explanation for Robustness}

\begin{figure}[h!]
    \centering
    \begin{subfigure}{\columnwidth}
    \centering        \includegraphics[width=0.48\textwidth,height=4cm]{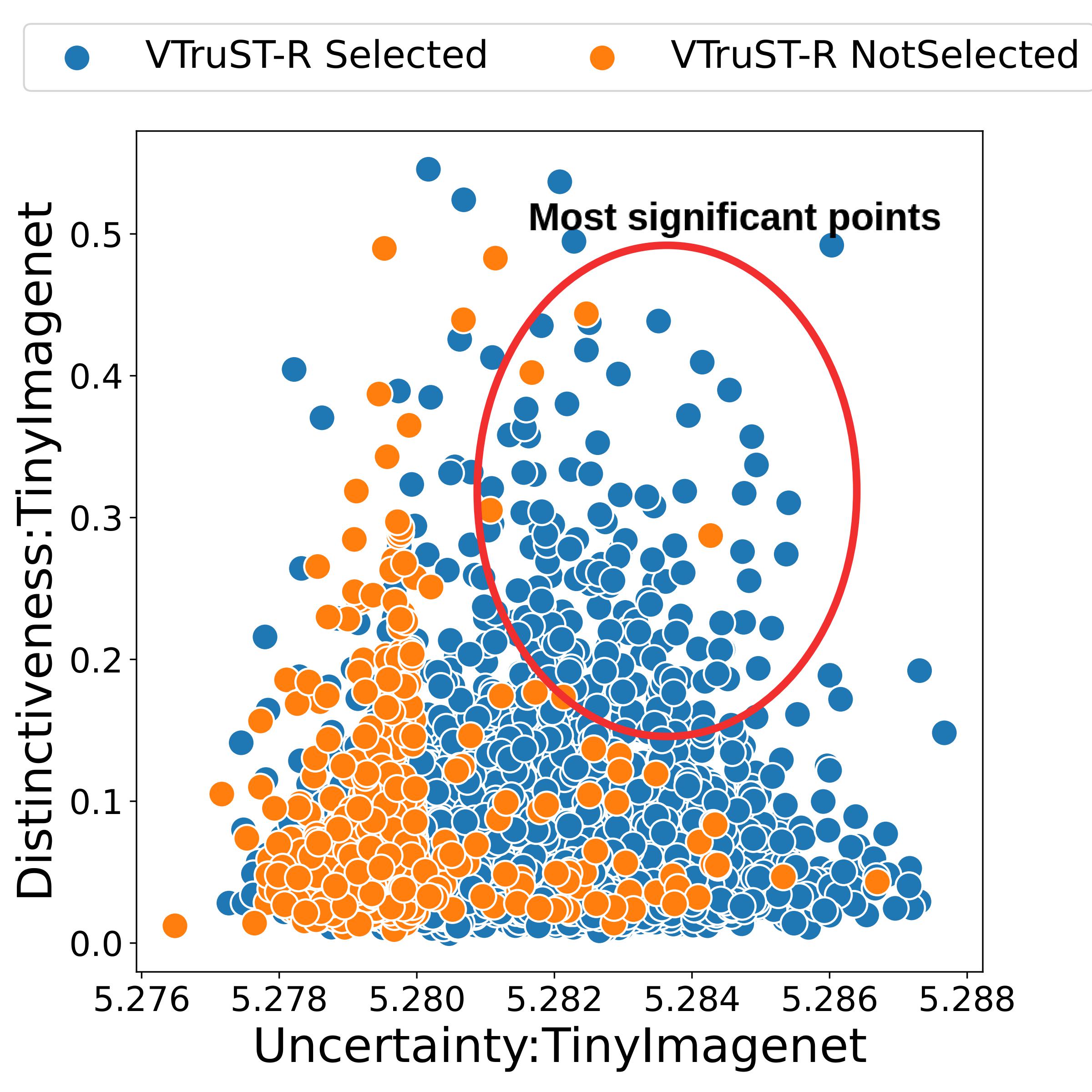}
    \end{subfigure}
    \begin{subfigure}{\columnwidth}
    \centering        \includegraphics[width=0.48\textwidth,height=4cm]{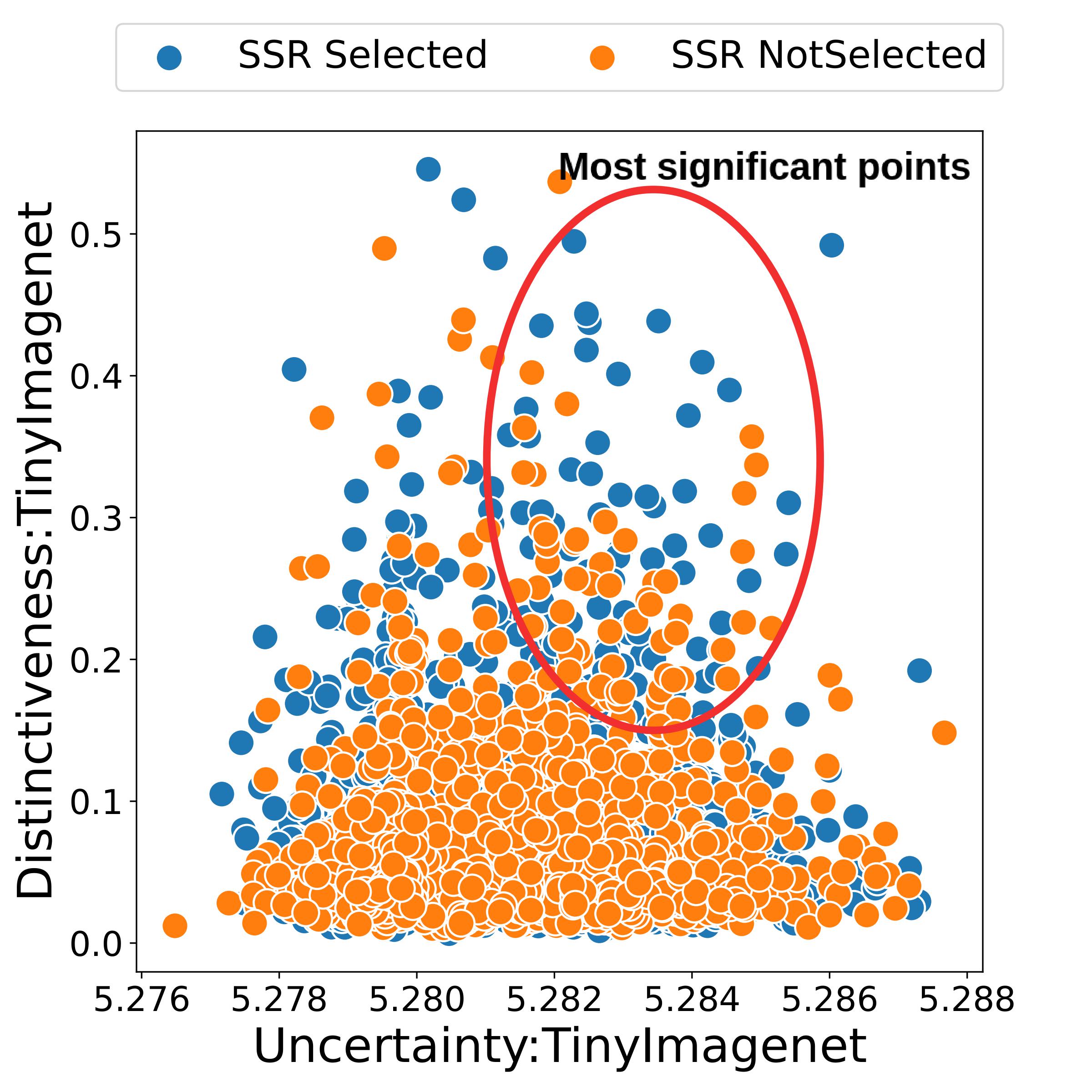}
    \end{subfigure}

    \begin{subfigure}{\columnwidth}
    \centering        \includegraphics[width=0.48\textwidth,height=4cm]{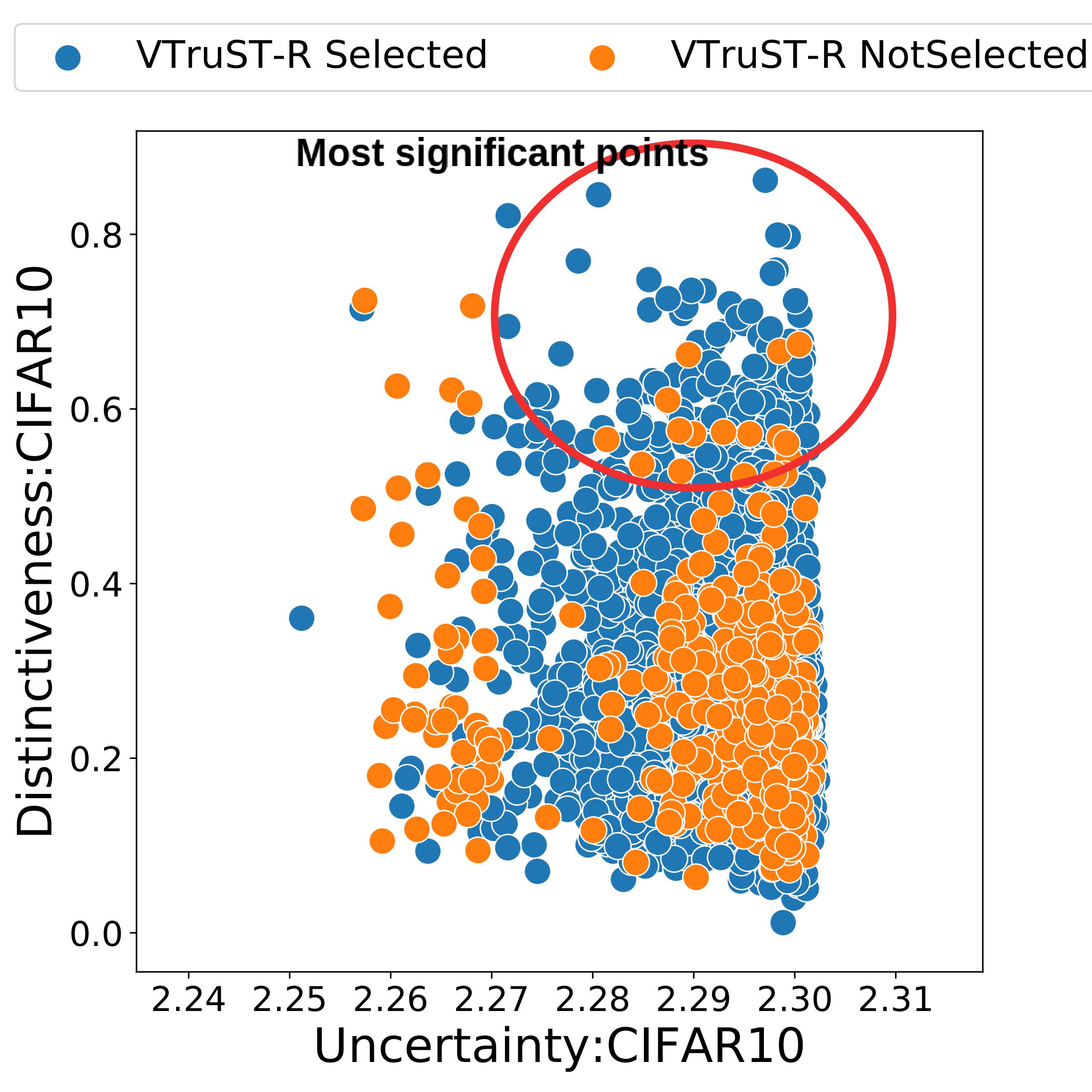}
    \end{subfigure}
    \begin{subfigure}{\columnwidth}
    \centering        \includegraphics[width=0.48\textwidth,height=4cm]{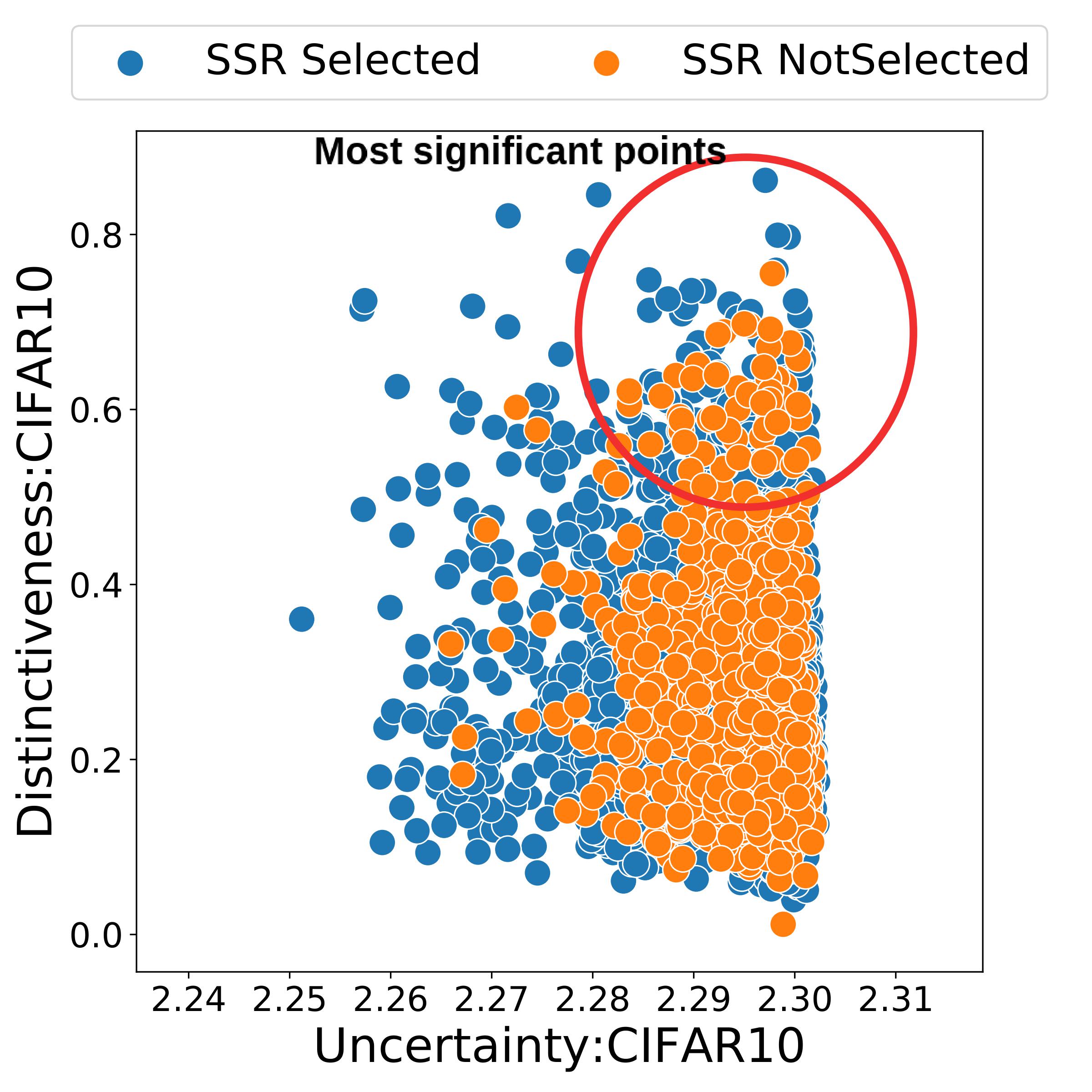}
    \end{subfigure}
    \captionsetup{font=scriptsize}
    \caption{\textbf{Data Map for randomly taken 5000 samples from TinyImagenet and CIFAR10 augmented training dataset }}
    \label{fig:datamap}
\end{figure}

We visualize the datapoints in the two dimensions - Uncertainty and Distinctiveness (defined in the main paper) in Figure \ref{fig:datamap} where we choose a random set of 5000 points from CIFAR and TinyImagenet datasets, followed by marking them as \textit{selected} and \textit{not selected} by VTruST-R and SSR respectively. We can observe that points with relatively high uncertainty and high distinctiveness(HD-HU) values mostly belong to the \textit{selected set of VTruST-R}, while the \textit{unselected points from SSR} mostly cover the HD-HU region.

\begin{figure}[h!]
    \centering
    \begin{subfigure}{\columnwidth}
    \includegraphics[width=0.8\textwidth,height=4cm]{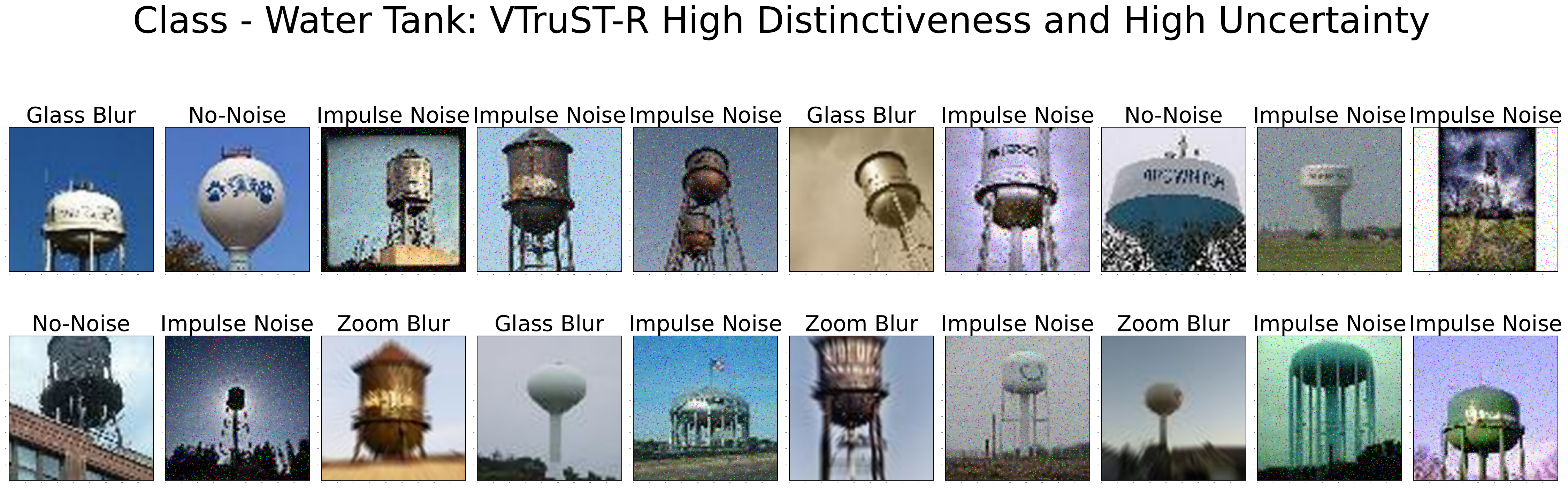}
    \end{subfigure}    

    \hspace{10mm}

    \begin{subfigure}{\columnwidth}
    \includegraphics[width=0.8\textwidth,height=4cm]{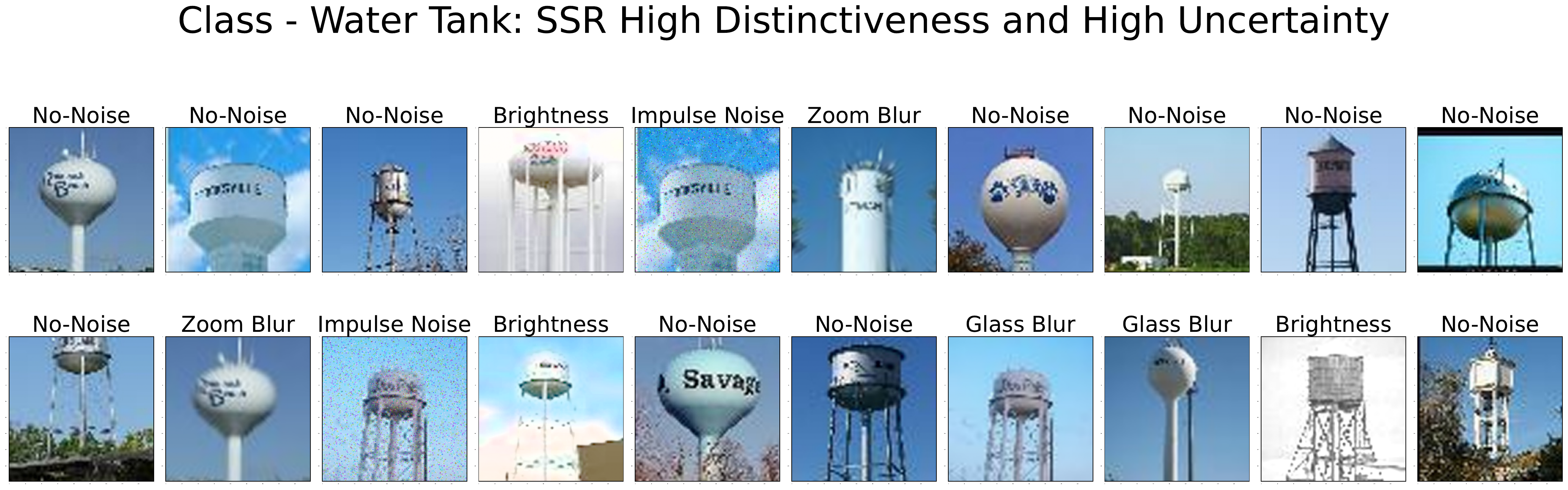}
    \end{subfigure}

    \caption{\textbf{Anecdotal samples from VTruST-R and SSR with High Distinctiveness-High Uncertainty from TinyImagenet for class Watertank.}}
    \label{fig:anec_high_low_tim_wc}
\end{figure}

\begin{figure}[h!]
    \centering
    \begin{subfigure}{\columnwidth}
    \includegraphics[width=0.8\textwidth,height=4cm]{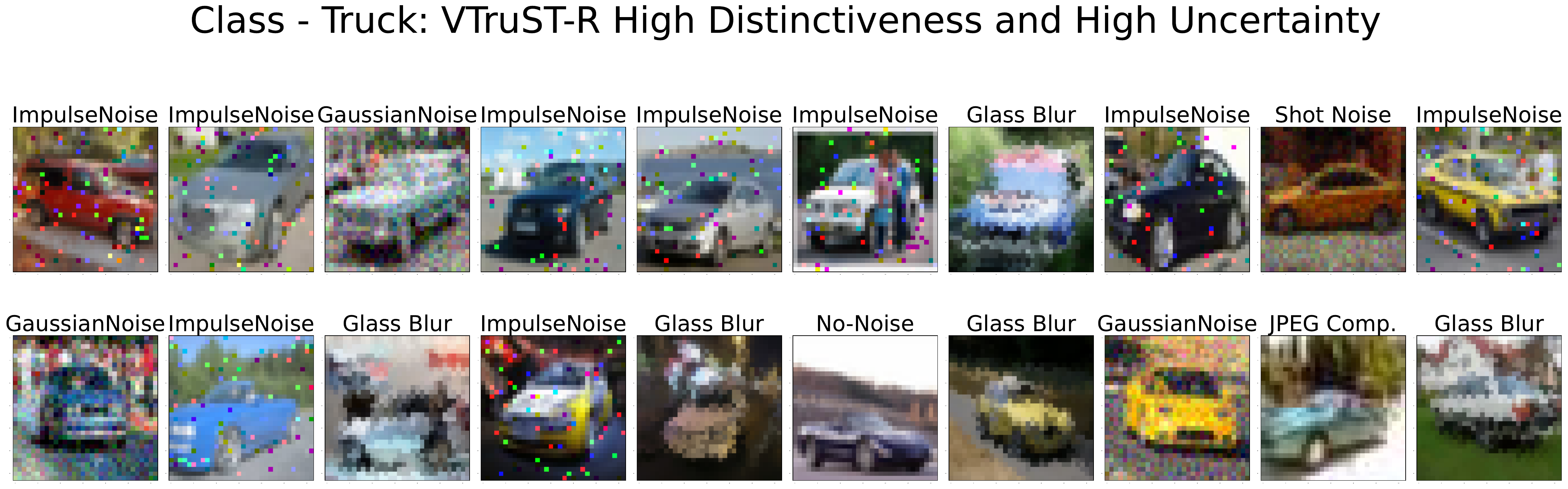}
    \end{subfigure}    

    \hspace{10mm}

    \begin{subfigure}{\columnwidth}
    \includegraphics[width=0.8\textwidth,height=4cm]{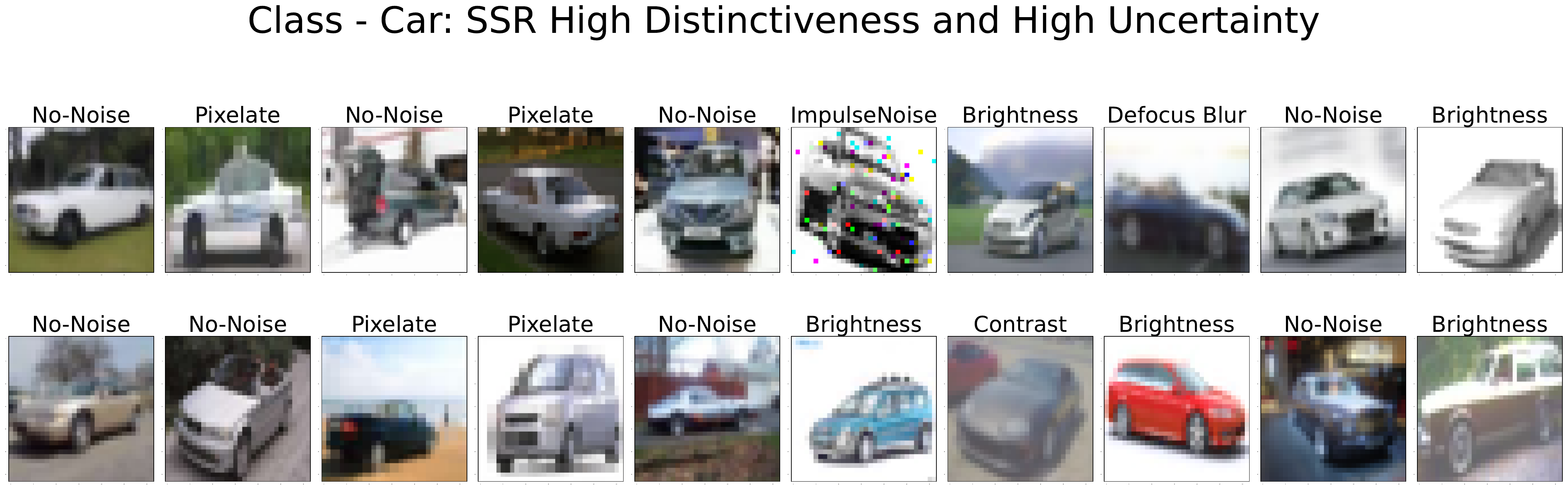}
    \end{subfigure}
    
    \hspace{10mm}
    
    \begin{subfigure}{\columnwidth}
    \includegraphics[width=0.8\textwidth,height=4cm]{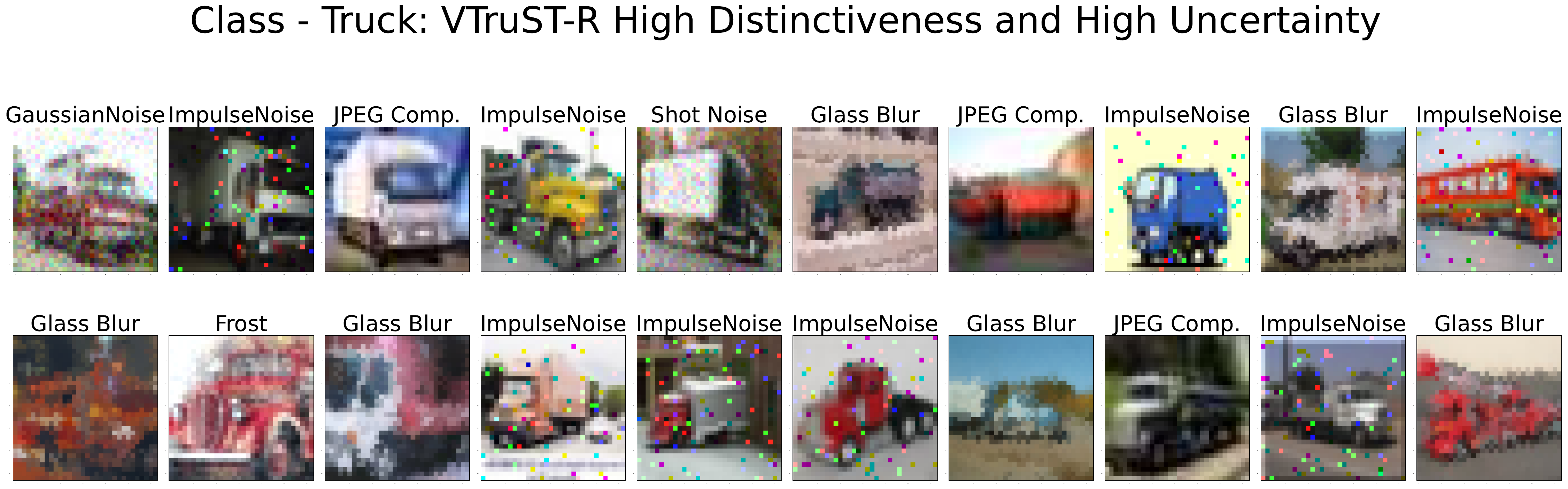}
    \end{subfigure}    

    \hspace{10mm}

    \begin{subfigure}{\columnwidth}
    \includegraphics[width=0.8\textwidth,height=4cm]{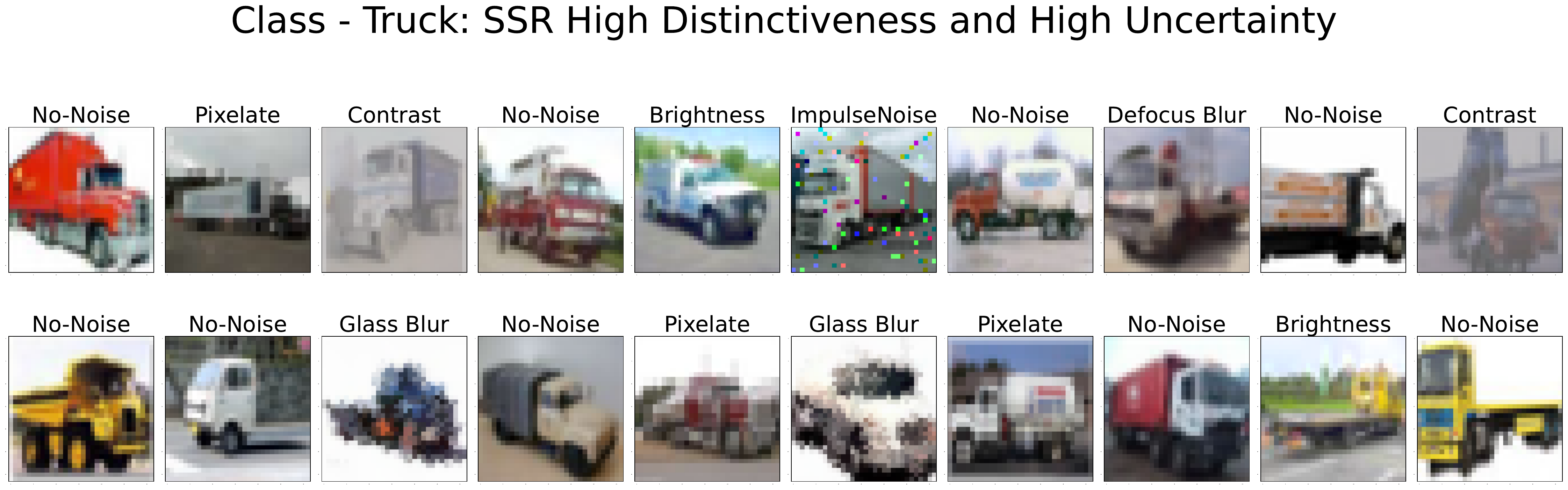}
    \end{subfigure}

    \caption{\textbf{Anecdotal samples from VTruST-R and SSR with High Distinctiveness-High Uncertainty from CIFAR10 for classes Car and Truck.}}
    \label{fig:anec_high_low_cifar}
\end{figure}

We show anecdotal samples for the class \textit{Watertank} from TinyImagenet in Figure \ref{fig:anec_high_low_tim_wc} and for the classes \textit{Car} and \textit{Truck} from CIFAR-10 in Figure \ref{fig:anec_high_low_cifar} having high distinctiveness and high uncertainty. It can be noted that (a) VTruST-R selects diverse samples while SSR selects similar (mostly similar background) samples ; (b) VTruST-R mostly selects samples from difficult augmentations like Impulse Noise and Glass Blur while SSR selects samples from unaugmented (No-Noise) or easier augmented samples like Brightness and Contrast. This justifies the outperforming capability in robustness from VTruST-R in comparison with SSR.

\section*{6.Details on training regime}

\textbf{Experiments using Social Data:} For all the datasets, we use a 2-layer neural network and vary the learning rate on a grid search between $5^{-1}$ to $5^{-4}$.
\newline
\textbf{Experiments using Image Data:} For all the datasets, we use ResNet-18 model and a learning rate of $10^{-1}$ with momentum of $0.9$ and weight decay of $5^{-4}$.
\newline
\textbf{Experiments using Scientific Data:} For all the datasets, we use convolutional neural networks as the experimental setup from \citep{Benato_2022} and vary the learning rate on a grid search between $10^{-2}$ to $10^{-4}$.

\end{document}